\documentclass[journal]{IEEEtran}
\ifCLASSINFOpdf
\else
\fi
\usepackage{url}

\usepackage[colorlinks]{hyperref}
\usepackage{graphicx}
\usepackage{booktabs}

\begin{document}
%
\title{KOLOMVERSE: Korea open large-scale image dataset for object detection in the maritime universe}
%
%
%

\author{Abhilasha Nanda,
        Sung Won Cho,~\IEEEmembership{Member,~IEEE,}
        Hyeopwoo Lee,
        and Jin Hyoung Park
\thanks{This research was supported by Korea Research Institute of Ships and Ocean engineering a grant from Endowment Project of “Development of Open Platform Technologies for Smart Maritime Safety and Industries” funded by Ministry of Oceans and Fisheries(2520000292, PES5230), and also supported by the National Information Society Agency of "Marine Object AI Data" funded by Ministry of Science and ICT. Corresponding authors: Sung Won Cho}

\thanks{Abhilasha Nanda, Hyeopwoo Lee and Jin Hyoung Park are with the AIVENAUTICS, Daejeon, Republic of Korea (e-mail: annie.nanda@aivenautics.com, hyeopwoo.lee@aivenautics.com, jinhyoung.park@aivenautics.com).}%
\thanks{Sung Won Cho is with the Department of Management Engineering at Dankook University in Cheonan, Republic of Korea (e-mail: sungwon.cho@dankook.ac.kr).}}

%
%

\markboth{Journal of \LaTeX\ Class Files,~Vol.~00, No.~00, August~2024}%
{Shell \MakeLowercase{\textit{et al.}}: Bare Demo of IEEEtran.cls for IEEE Journals}
%



\maketitle

\begin{abstract}
 Object detection in the maritime domain is crucial for ensuring the safety and navigation of ships. However, there is still a lack of publicly available large-scale datasets in this domain. To address this challenge, we introduce KOLOMVERSE, an open large-scale image dataset for object detection in the maritime domain. We gathered 5,845 hours of video data captured from 21 territorial waters of South Korea. Through rigorous data quality assessment, we manually gathered around 186,419 4K resolution images from the video data. The KOLOMVERSE includes five classes (ship, buoy, fishnet buoy, lighthouse and wind farm) for maritime object detection. The dataset comprises images with dimensions of 3840$\times$2160 pixels and, to the best of our knowledge, it is by far the largest publicly available dataset for object detection in the maritime domain. We conducted object detection experiments and evaluated our dataset using several state-of-the-art pre-trained architectures to demonstrate its effectiveness and usefulness. The dataset is available at: \url{https://github.com/MaritimeDataset/KOLOMVERSE}.
\end{abstract}


%
\IEEEpeerreviewmaketitle

\section{Introduction} \label{section 1}
%
%
%
%
\IEEEPARstart{I}{n} the field of computer vision and image processing, object detection is immensely popular because of its ability to accurately identify and classify objects from images. It plays a crucial role in localizing and tracking objects in scenes. Due to its extensive application, there are many publicly available datasets for implementation, such as MS COCO \cite{lin2014microsoft}, Pascal VOC \cite{everingham2010pascal}, Imagenet \cite{deng2009imagenet} and Google Open Images \cite{krasin2017openimages} to name a few. MS COCO, which contains 328K images across 80 object categories, is the most frequently used public dataset. Object detection methods have been applied to tasks such as face identification \cite{hjelmaas2001face}, pedestrian detection \cite{dollar2009pedestrian}, vehicle position and velocity estimation \cite{friedland1973optimum} and autonomous vehicles \cite{wen2021fast}. These deep learning techniques have become popular in the maritime domain as well with implementation of maritime object detection, segmentation, and vessel re-identification among others. However, collecting and annotating data for these applications is challenging due to the complexity and uncertainty of the sea environment \cite{wang2020ocean}. Therefore, there are few maritime datasets and and most of them are not publicly accessible.

With the expansion of the shipping industry worldwide, maritime datasets play a pivotal role in the development of Maritime Autonomous Surface Ships (MASS) as well as navigation safety \cite{elkins2010autonomous,liu2020data}. These datasets encompass information about vessels and surrounding water objects, enabling the detection, monitoring, and management of sea traffic, maritime safety, navigation, and military surveillance to prevent illegal activities such as smuggling, infiltration, and dumping of pollutants.

To ensure safety both during the day and night and prevent accidents, maritime image datasets provide information such as a ship’s location, size, and direction, crucial for identifying nearby objects\cite{shao2018seaships}. While object detection applications like harbor surveillance \cite{ray2019heterogeneous} and collision avoidance \cite{ramos2019collision} for autonomous vessels exist, the maritime domain faces challenges due to a scarcity of publicly available data and the complexities of the sea environment, which make it an expensive field of research\cite{wang2020ocean}. Unlike other object detection domains such as street signs detection, \cite{piccioli1996robust}, pedestrian detection \cite{dollar2009pedestrian}, or even face detection \cite{hjelmaas2001face}, maritime object detection lacks comparable public benchmarks, resulting in limited studies and results.

In this paper, we introduce KOLOMVERSE, a maritime object image dataset designed specifically for object detection. For data collection, we deployed a four-channel 4K (3840$\times$2160 pixels) Network Video Recorder (NVR) system on ships' front, rear, left, and right sides to capture video data at one frame per second (fps), from which KOLOMVERSE was constructed using extracted images. Seventeen ships equipped with the NVR system collected the video data across 21 territorial waters of Korea, mapped onto Google Earth satellite imagery as depicted in Figure~\ref{Fig1}. From November 2020 to February 2021, we accumulated 5,845 hours of video data. Through rigorous data quality assessment, we curated approximately 186,419 images in 4K resolution, representing diverse environmental conditions such as weather, time (illumination and occlusion), viewpoint, background, wind speed, visibility, and seasons. The KOLOMVERSE contains five categories (ship, buoy, fishnet buoy, lighthouse, and wind farm) essential for maritime vessel safety and navigation.

Numerous object detection models are publicly available including the YOLO series \cite{redmon2016you, redmon2017yolo9000, redmon2018yolov3, bochkovskiy2020yolov4}, SSD \cite{liu2016ssd}, fast and faster RCNN (FRCNN) \cite{girshick2015fast, ren2015faster}, R-FCN \cite{dai2016r}, Mask R-CNN \cite{he2017mask} and CenterNet \cite{zhou2019objects}. We trained and evaluated our dataset using pre-existing state-of-the-art models to demonstrate its effectiveness in deep learning applications and its potential for broader applications. We trained our dataset on benchmarking models, so the results obtained from our dataset are reproducible. We developed KOLOMVERSE to pave the way for future advancements in maritime object detection. Our dataset and the trained models are publicly available at \url{https://github.com/MaritimeDataset/KOLOMVERSE}.

\section{Related works}\label{section 2}

There are a handful of maritime datasets available for ship detection and classification. Prasad et al. \cite{prasad2017video} introduced the Singapore Maritime Dataset (SMD), which includes 10 categories of images: Ferry, buoy, vessel/ship, speed
boat, boat, kayak, sail boat, swimming
person, flying bird/plane, and other. It is one of the few publicly available datasets dedicated specifically to object detection in maritime environments, but it lacks representative benchmarking results. Zhang et al. \cite{zhang2015vais} introduced Vais, a large dataset that containing over 1,000 paired RGB and infrared images across six ship categories: merchant, sailing, passenger, medium, tug, and small. The dataset comprises 2,865 images, including 1,623 visible, 1,242 IR, and 154 nighttime IR images. It features 264 uniquely named ships into six coarse-grained categories: merchant ships, sailing ships, medium passenger ships, medium "other" ships, tugboats, and small boats.
Shao et al. \cite{shao2018seaships} developed Seaships for training and evaluating ship object detection algorithms, containing 31,455 images across six ship classes: ore carrier, bulk cargo carrier, general cargo ship, container ship, fishing boat, and passenger ship. Soloviev et al. \cite{soloviev2020comparing} curated 135 videos from a sightseeing watercraft operating between the cities of Turku and Ruissalo in South-West Finland, along the Aura river and into the Finnish Archipelago. This dataset includes 850 annotated vessels. They also built a second real maritime dataset in the Finnish Archipelago, featuring 1,750 images captured with visible cameras in open sea landscapes, comprising a total of 9,137 vessel objects.

Other maritime datasets address challenges such as piracy detection, boat monitoring, segmentation, and vehicle classification. The IPATCH dataset (piracy detection) developed by Pationo et al. \cite{patino2016pets}, focuses on piracy detection using multi-sensor surveillance to protect vessels at sea. The dataset adresses challenges from the PETS 2016 workshop, does low-level video analysis such as object detection and tracking, mid-level video analysis such as 'simple' event detection (the behavior recognition of a single actor), and high-level video analysis that is 'complex' event detection (behavior and interaction recognition of several actors). The Maritime Detection, Classification, and Tracking Database (MarDCT) \cite{bloisi2015argos} is a repository of videos and images sourced from various cameras (fixed, moving, and Pan-Tilt-Zoom) and scenarios. MarDCT aims to provide visual data to support the development of intelligent surveillance systems for maritime environments, categorized by ground truth types: detection, classification, and tracking. Also known as the Argos Boat Classification dataset, it focuses on boat classification using images automatically extracted by the ARGOS system, operating continuously in Venice, Italy. The dataset comes from an incomparable environment like Venice, but they present very interesting challenges to vehicle classification. The Marine Obstacle Detection Dataset (MODD), introduced by Kristan et al. \cite{kristan2015fast}, consists of marine videos captured by unmanned surface vehicles (USVs). This dataset facilitates image segmentation into the sky, the shore, and sea areas, focusing on obstacle detection within the marine environment.

The contributions of our study compared with those of the aforementioned studies are summarized as follows:

\begin{enumerate}
  \item Autonomous navigation systems rely on comprehensive environmental information for safe operation. Our dataset encompasses a diverse range of maritime objects and environmental features, such as fishnet buoys, lighthouses, and wind farms, collected from 21 different water bodies. This diversity in background, scale, viewpoint, occlusion, and illumination ensures that models trained on our dataset are robust and reliable for real-world applications.
  \item KOLOMVERSE is significantly larger than other maritime datasets. Datasets with fewer samples are typically less informative and prone to overfitting, resulting in biased and unreliable models \cite{shorten2019survey}. In the realm of maritime applications, the need for dependable models that guarantee safety is crucial.
  \item Recognizing even the smallest objects from a distance is crucial for safe navigation in real-time sea environments. KOLOMVERSE is constructed using high-resolution images (3840$\times$2160 pixels) that capture various shapes and sizes, providing detailed information necessary for detecting and classifying small and distant objects.
\end{enumerate}

\begin{table*}[ht]
  \caption{Comparison of the KOLOMVERSE with publicly available maritime object detection datasets.}
  \label{Tbl1}
  \centering
  \begin{tabular}{llllp{3cm}p{5cm}}
    \toprule
    Datasets  & $\#$ of images & $\#$ of classes & Image size & Diversity & Applicability\\
    \midrule
    Seaships \cite{shao2018seaships} & 31,455 & 6  & 1920$\times$1080& only maritime vessels & Vessel detection and classification\\
    VAIS \cite{zhang2015vais} & 2,865 & 6  & 1024$\times$768&only maritime vessels & Vessel detection and classification \\
    SMD \cite{prasad2017video} & 17,450 & 10  & 1920$\times$1080& various maritime vessels, swimming person,  flying bird/plane, and other& Ship, swimming person and bird detection but lacks important targets for safe maritime navigation\\
    MarDCT \cite{bloisi2015argos}&8,115&25&377$\times$188&24 boat categories and water &Maritime vehicle detection, classification and tracking, surveillance applications\\
    \cmidrule(r){1-6}
    KOLOMVERSE & 186,419  & 5 & 3840$\times$2160&Ship, buoy, fishnet buoy, lighthouse, and wind farm& Diverse important maritime object detection for real time safe navigation in the sea\\     
    \bottomrule
  \end{tabular}
\end{table*}

\section{Dataset}\label{section 3}

\subsection{Data collection and preprocessing}

\begin{figure}[b!]
\centering
\includegraphics[width=0.8\linewidth]{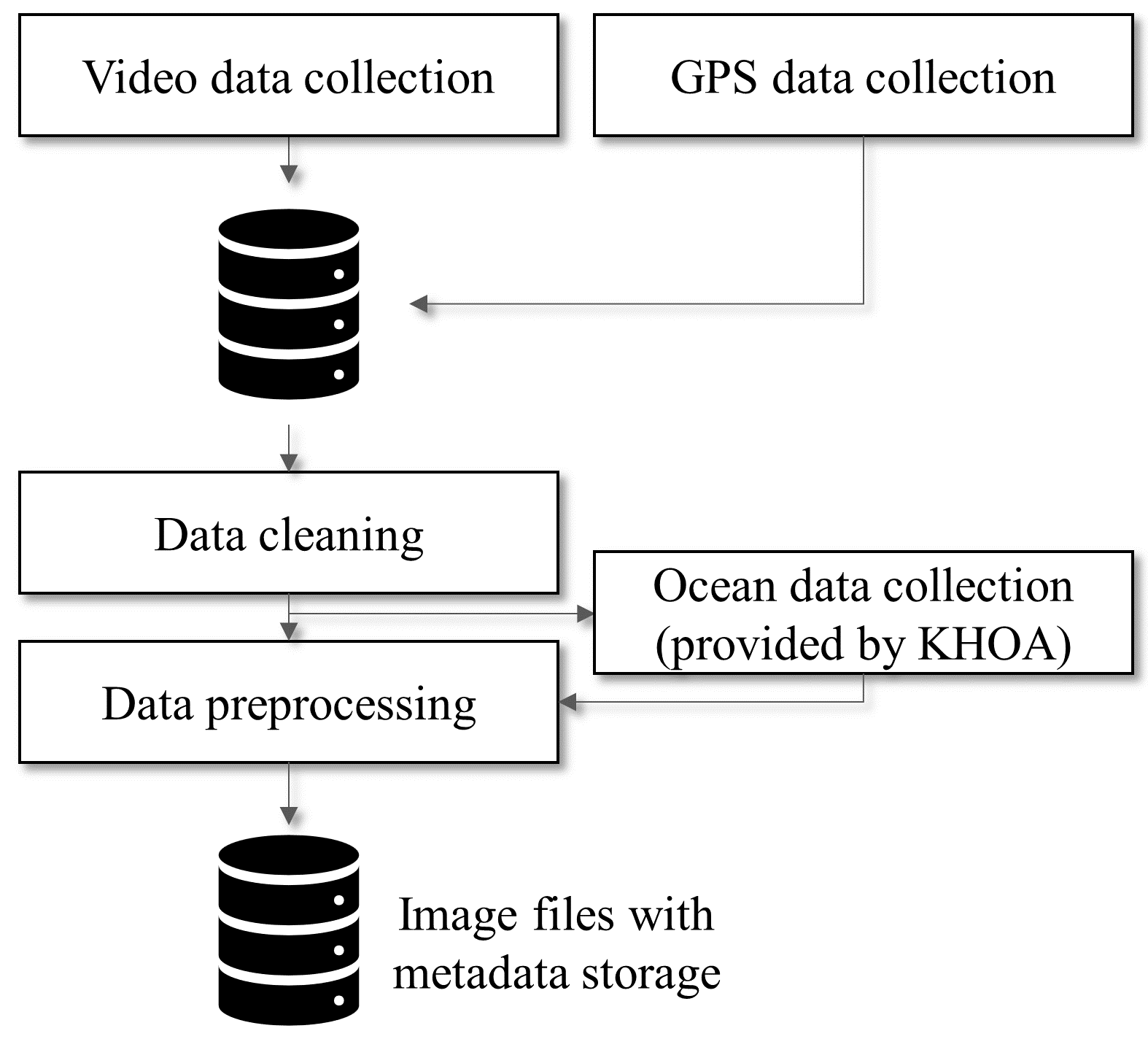}
\caption{Framework of data collection procedure.}
\label{Fig:data_process}
\end{figure}

\begin{figure}[ht]
 \centering
\includegraphics[width=0.385\linewidth]{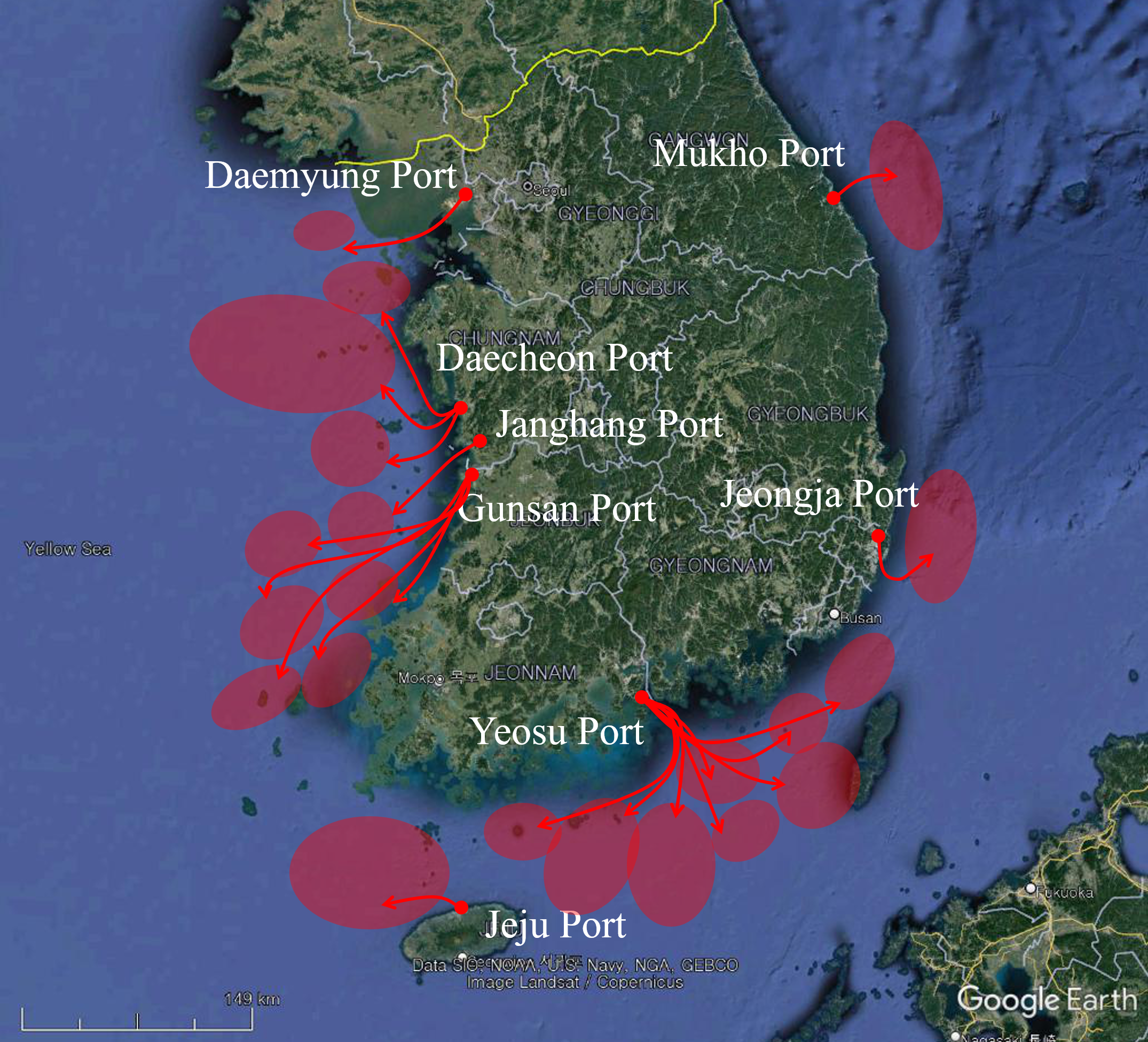}
\includegraphics[width=0.597\linewidth]{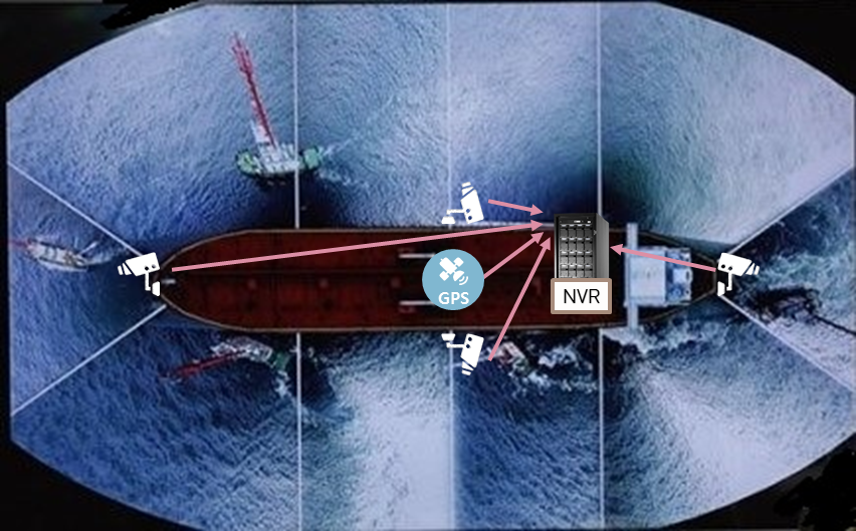}
\caption{\textbf{Left:} Twenty-one territorial waters of Korea for data collection. \textbf{Right:} Top view of the ship installed with four-channel 4K NVR system.}
 \label{Fig1}
\end{figure}

KOLOMVERSE was built through the procedure in Figure~\ref{Fig:data_process}. To obtain the video data, we installed a four-channel 4K NVR system as shown in Figure~\ref{Fig1}. We installed four cameras per ship to record the front, rear, left, and right directions along with GPS equipment to store location information simultaneously in the recorded image. To reflect the characteristics of the sea in Korea, we divided it into nine zones in the South Sea, 10 zones in the West Sea, and two zones in the East Sea to collect video data from various environments. The video data included variations in contrast and color of objects according to various weather conditions and times. 

\begin{figure}[ht]
 \centering
\includegraphics[width=0.49\linewidth]{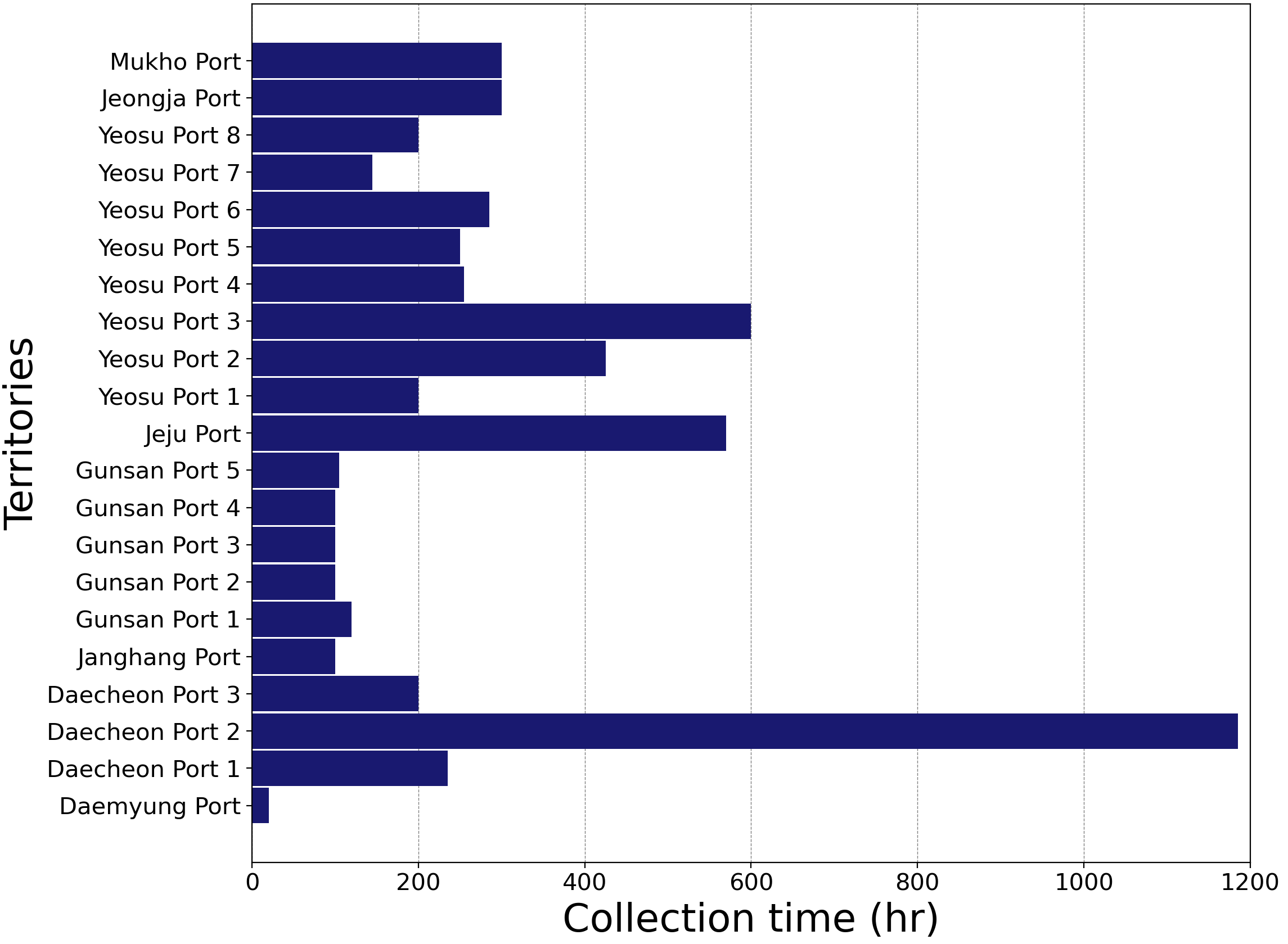}
\includegraphics[width=0.49\linewidth]{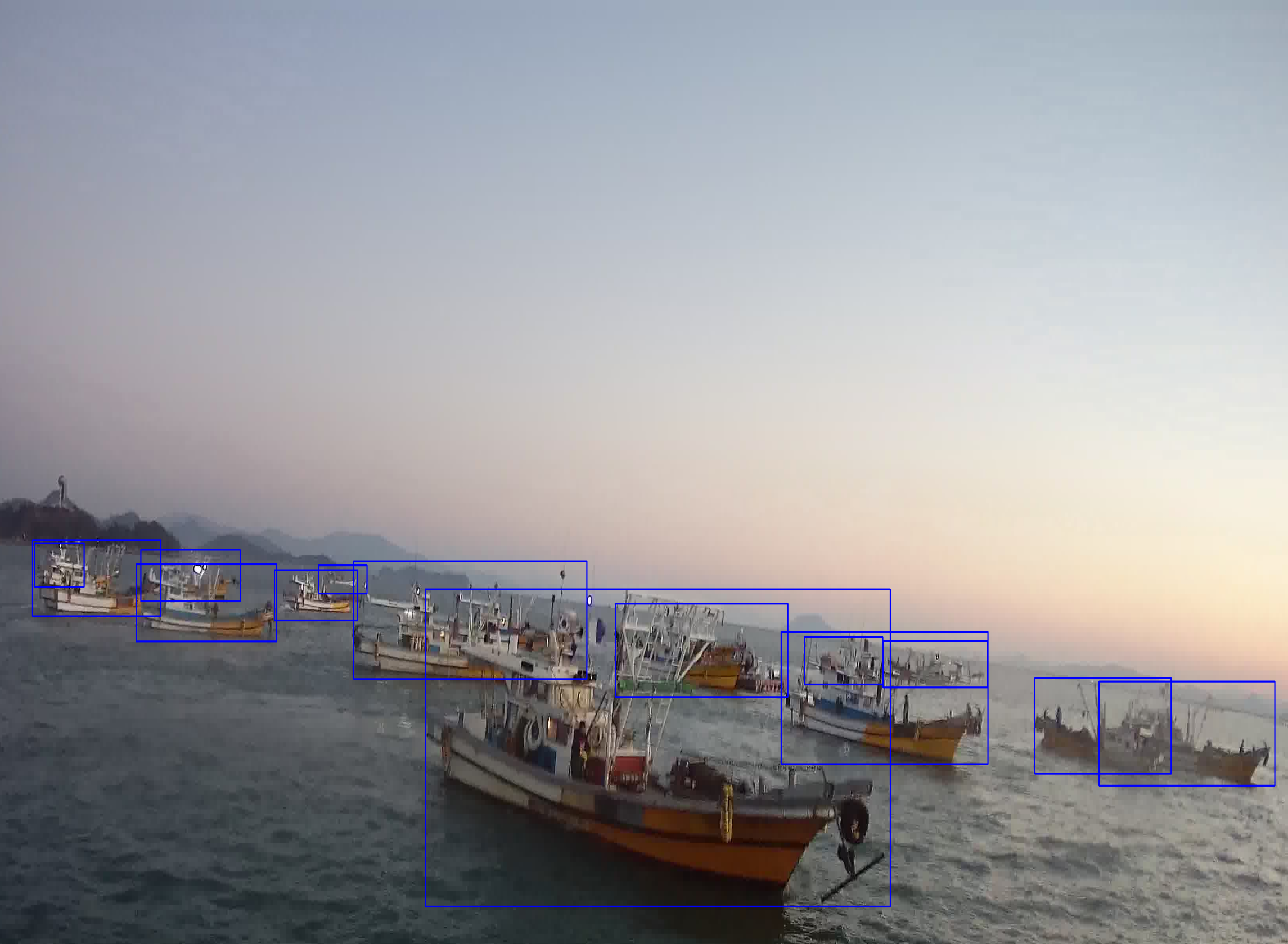}
\caption{\textbf{Left:} The distribution of the amount of time taken to collect the video data from 21 territories. \textbf{Right:} Image sample with partially visible and annotated objects.}
 \label{Fig2}
\end{figure}

We utilized 17 ships to gather video data, which was stored in mp4 format, totaling 5,845 hours of footage at 4K resolution. Figure~\ref{Fig2} shows the distribution of the amount of time taken to collect the video data from all the 21 territories.

The collected data underwent cleaning processes to ensure alignment between video frame timestamps and GPS data timestamps, while also removing instances with missing or outlier GPS data. Additionally, utilizing GPS data, we retrieved metadata for ocean information (wave height, wind speed, and visibility) provided by the Korea Hydrographic and Oceanographic Agency (KHOA) \cite{KHOA}. Data were collected according to specific criteria as follows:

\begin{itemize}

  \item Weather: The videos were collected by categorizing weather conditions into clear, cloudy, rainy, snowy, and strong windy. At wind speeds of 14 m/s or more per second, sailing is prohibited, so such data were excluded from the collection.
  \item Time: The color, clarity, and contrast of objects vary with time due to changes in illumination and direction of light in the images. To account for this variability, we classified the videos into categories such as morning, afternoon, sunrise, sunset, and night.
  \item Wave height: Since the shaking and inclination of the images and the shapes of the objects in the video vary depending on the wave height, we collected the video by classifying it into 0, 1 to 2 meters, and 3 to 4 meters. Sailing is prohibited for the wave height of 4 meters or more, so those data were excluded from the collection. 
  \item Wind speed: As the movement of objects in the image and the shape of waves differ depending on the wind speed, the wind speed information was classified as 0, 1 m/s to 5 m/s, 5 m/s to 10 m/s, and 10 m/s or more. Sailing is prohibited at wind speeds of 14 m/s or higher, so such data were excluded from the collection.
  \item Visibility: Since the color, resolution, and clarity of objects in the video vary depending on the visibility, we collected the video by classifying 0.1 to 0.2 km, 0.2 to 0.5 km, 0.5 to 1 km, 1 to 2 km, and 2 km or more.
  
\end{itemize}

The collected video data were converted to image data in a JPG format. Image sequences were captured at one frame per second (1 fps), resulting in some redundancy. To minimize this redundancy, we removed images with overlapping areas of more than 80$\%$ between consecutive frames. Furthermore, the extracted video frames were integrated with metadata, including GPS and ocean information, and a database was built in a structured format that links each image to its corresponding metadata.

\subsection{Data annotation}

The image data were annotated into five classes: ship, buoy, fishnet buoy, lighthouse and wind farm. Initially, images that consisted solely of sea and islands without any objects from these five classes were removed. Additionally, images with sunlight glare over an object, potentially obscuring the object, or images of poor quality, or insufficient lighting hindering object identification were excluded from annotation.

Furthermore, annotated bounding boxes were required to encompass all parts of the object, with gaps between the object and bounding box less than 20 pixels. For ship classes, instances where persons or other objects were captured within a ship were collectively labeled as part of the ship. Additionally, objects partially visible due to occlusion or truncation were annotated as depicted in Figure~\ref{Fig2}. 

Following these annotation guidelines, images were manually labeled with bounding boxes, and each bounding box was recorded in a CSV file. A metadata file was also created for each image in the dataset, adhering to standard dataset practices.


\subsection{Data quality assessment}

\begin{figure}[ht]
\centering
\includegraphics[width=0.435\linewidth]{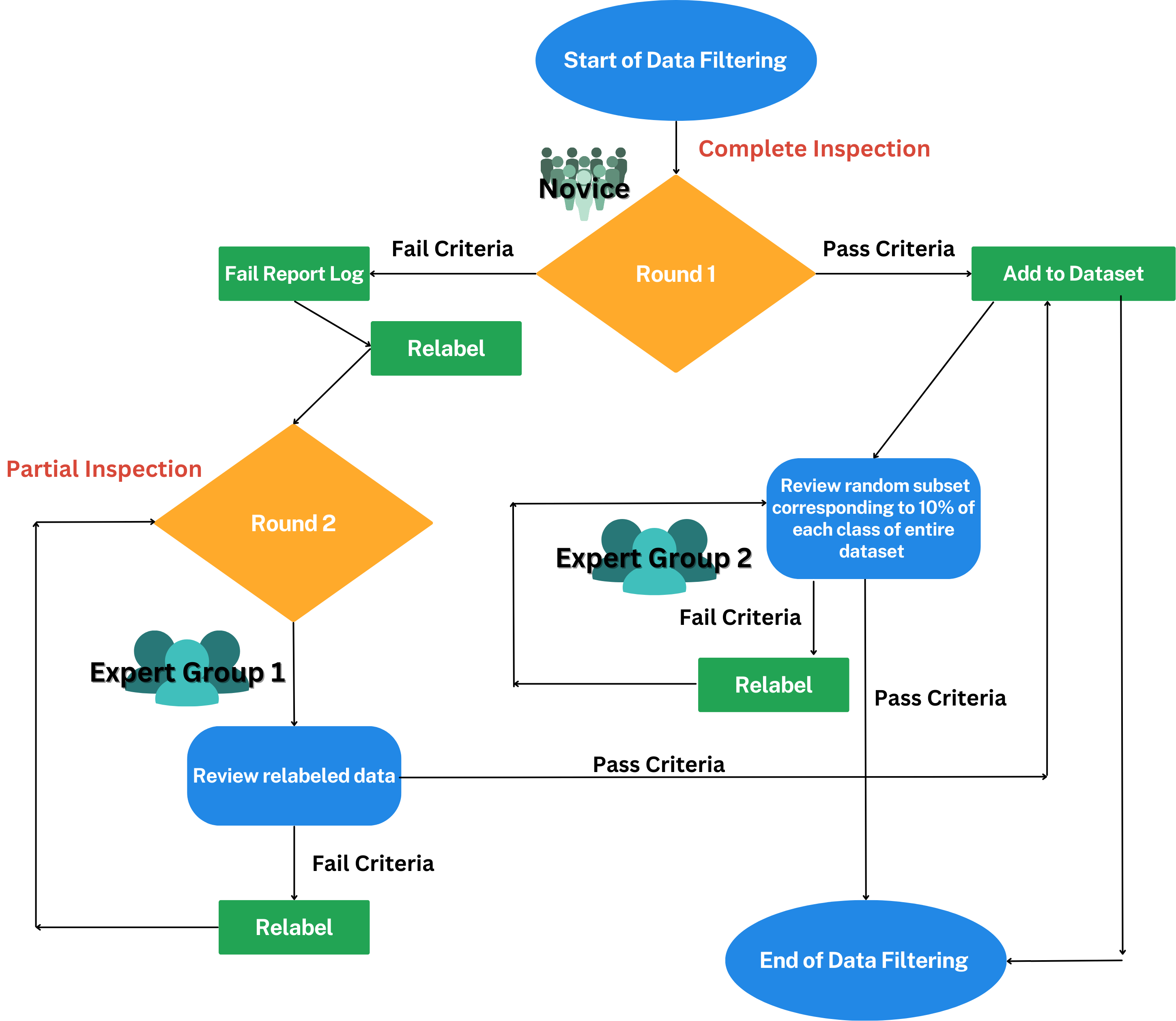}
\includegraphics[width=0.545\linewidth]{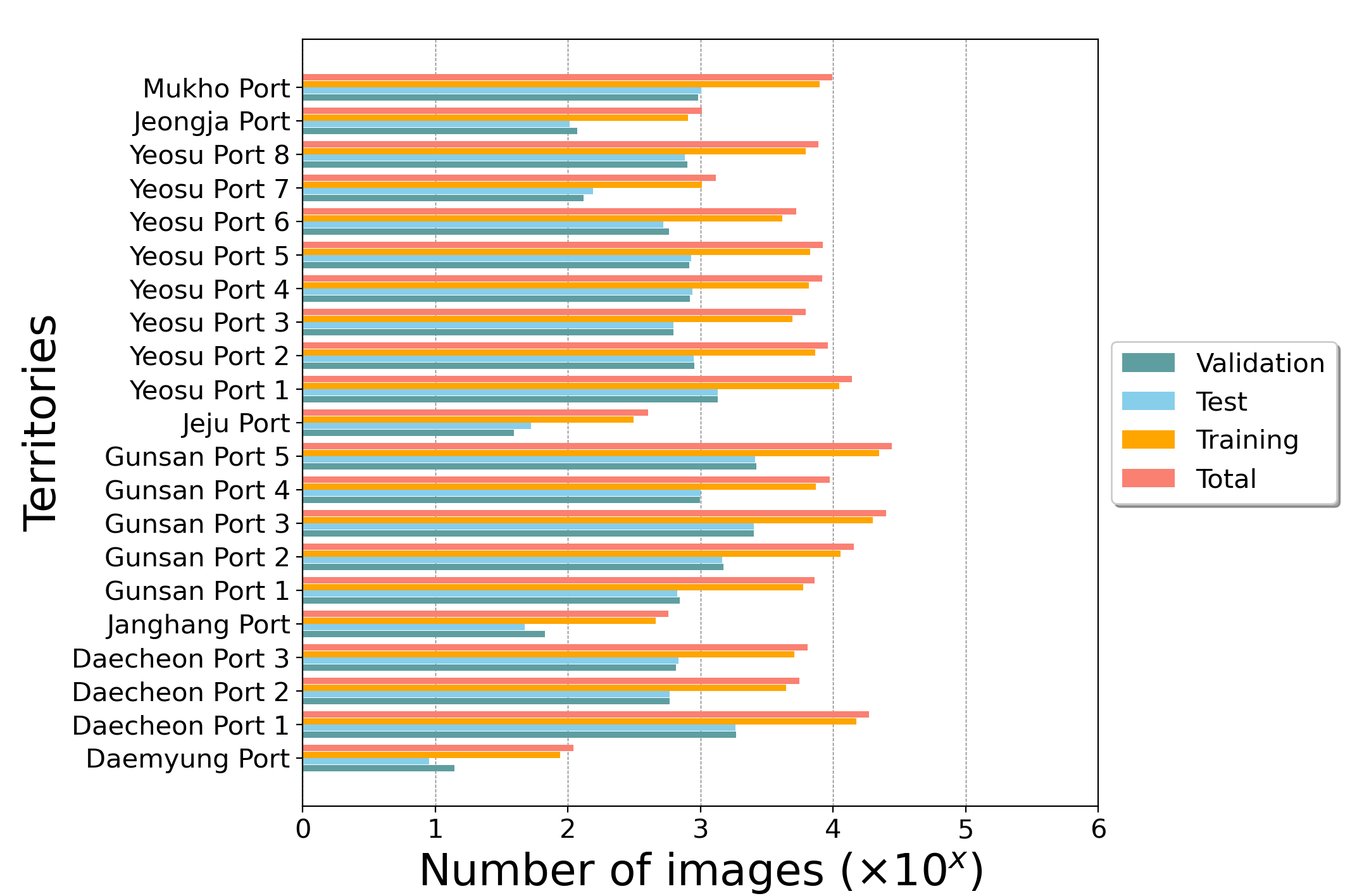}
\caption{\textbf{Left:} Framework of data quality assessment procedure. \textbf{Right:} Distribution of the train-validation-test split in the 21 territories.}
\label{Fig3}
\end{figure}

For post-annotation, we investigated our dataset to identify errors and correct them to form an accurate and reliable dataset. A two-step procedure to improve the data quality and reduce human errors during annotation processing, as shown in Figure~\ref{Fig3}.

In Round 1, novice annotators checked all the following anomalies through a thorough examination of the data. They checked for these errors and, if found, re-labeled the image correctly and delivered it to round 2.

\begin{itemize}

  \item Field validation: we checked that all necessary fields (e.g., image ID, class label, coordinate information) were presented in the annotation data.
  \item Error detection: we checked if the annotation criteria as described above were clearly followed. Moreover, we checked if any objects were incorrectly labeled and any objects that did not belong to any of our defined five classes were labeled.
  \item Class consistency: we ensured that objects of the same class are labeled consistently. Additionally, if the parts of the own ships used to collect our data are visible in the dataset, such ships were excluded from annotation as they have completely different shapes from the rest of the ships in the dataset.
  \item Coordinate consistency: we checked if the bounding box is exactly positioned on the object.
  
\end{itemize}

In Round 2, two groups of expert annotators were tasked with examining the data. The first group reviewed all the data that had been re-labeled in Round 1. Then, the second group selected a random subset corresponding to 10$\%$ for each class from the entire dataset, including the data reviewed by the first group, and thoroughly reviewed the selected samples to evaluate the quality of the annotations.

As a result, from the original video data, we acquired around 186,419 4K resolution images in JPG format to generate our dataset. The total 186,419 annotated images are divided into three parts, train (149,175 images), validation (18,643 images) and test (18,601 images) sets. Figure~\ref{Fig3} shows the data split in each of the 21 environments in our final dataset.

\section {Dataset description}\label{sec4}

\begin{figure}[ht]
  \centering
  \includegraphics[width=0.30\linewidth]{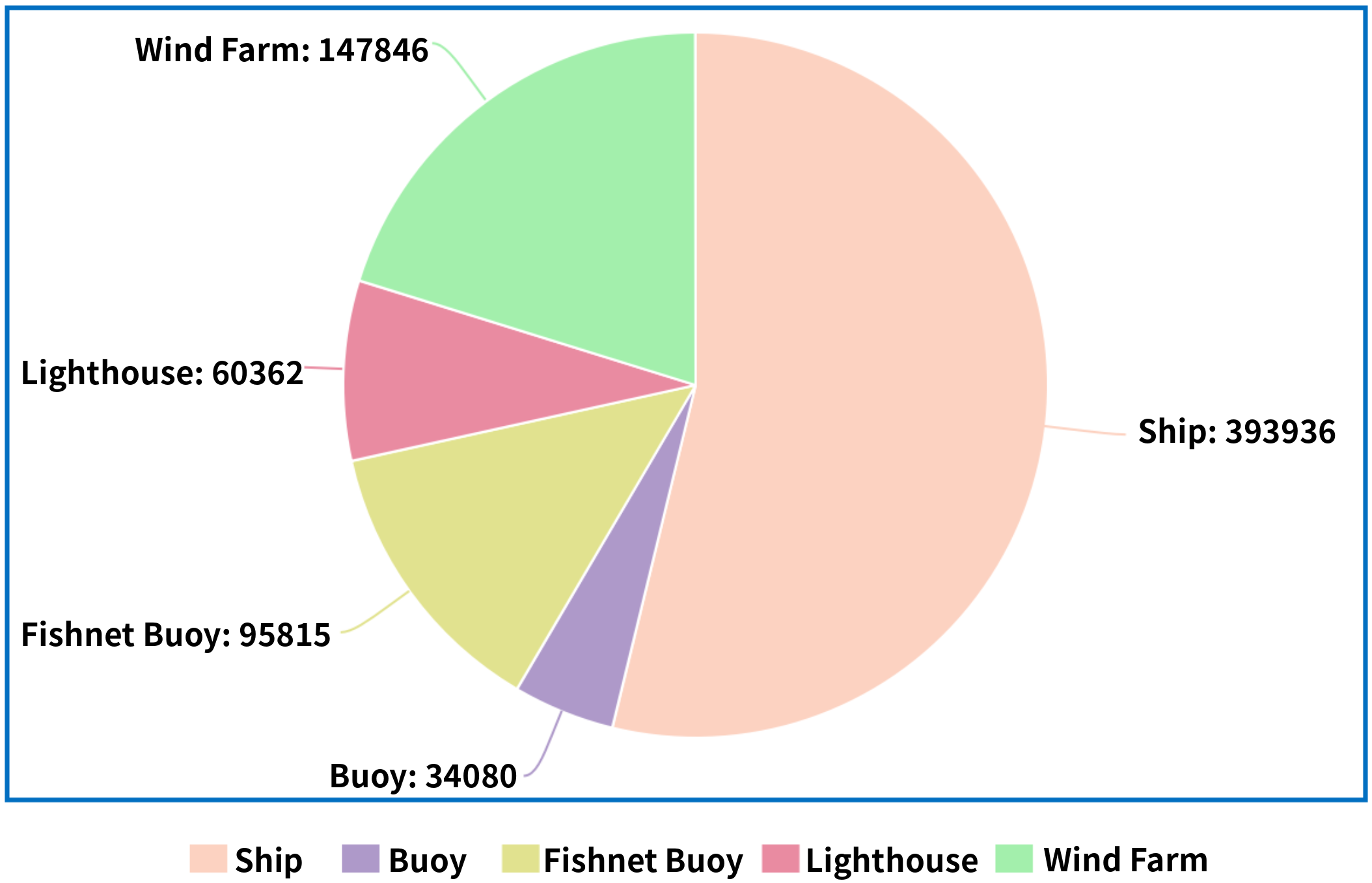}
  \includegraphics[width=0.685\linewidth]{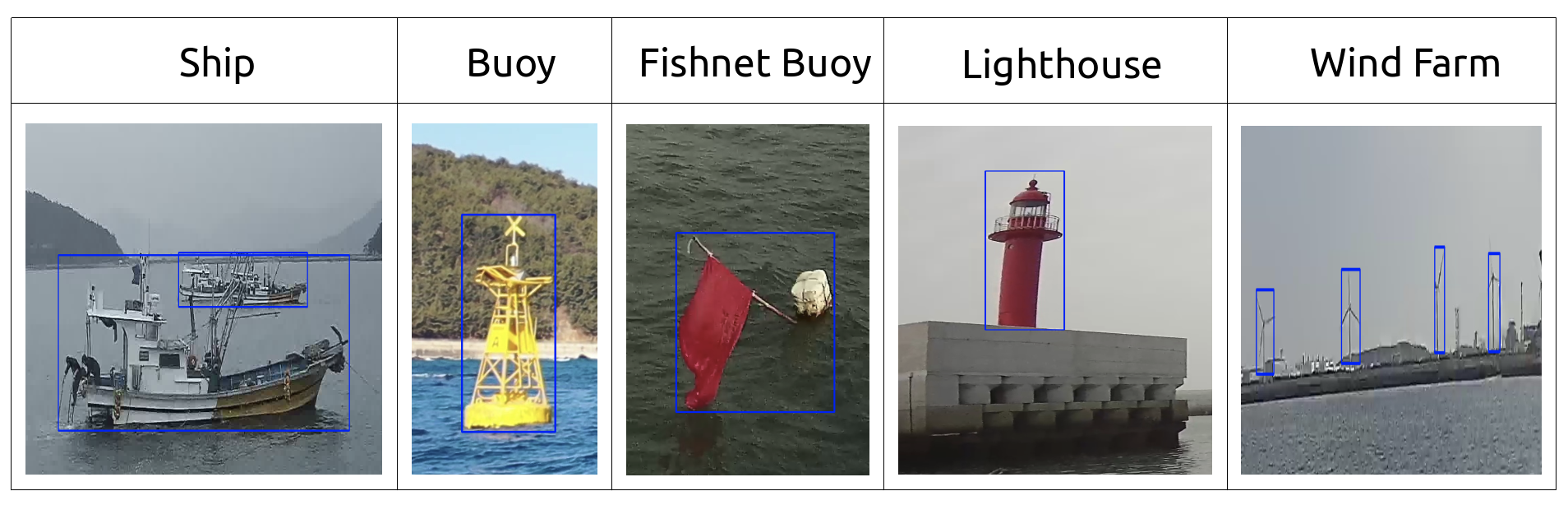}
  \caption{\textbf{Left:} Pie chart of class instances in the KOLOMVERSE. \textbf{Right:} Image samples containing objects from the five classes (ship, buoy, fishnet buoy, lighthouse and wind farm).}
   \label{Fig4}
\end{figure}

The KOLOMVERSE has a total of 186,419 4K images and contains images categorized into 5 distinct classes, namely ship (393,936 instances), buoy (34,080 instances), fishnet buoy (95,815 instances), lighthouse (60,362 instances) and wind farm (147,846 instances). Class-wise representation of the number of objects in the dataset is shown in Figure~\ref{Fig4}. It also shows image samples containing objects from the five classes. We present a dataset that can be used to develop robust and reliable deep learning models for real-time use. 

\begin{figure}[ht]
  \centering
  \includegraphics[width=0.49\linewidth]{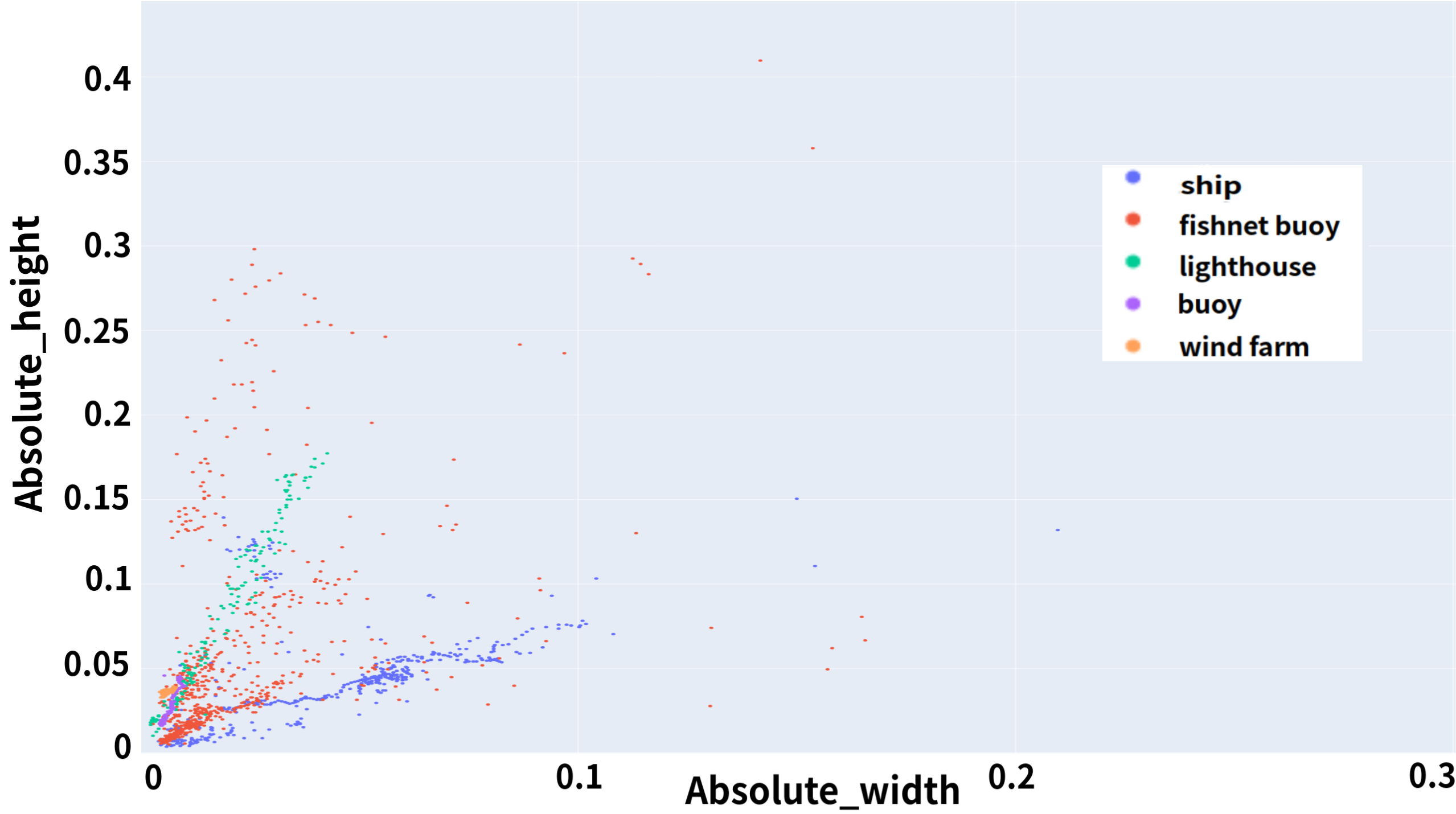}
  \includegraphics[width=0.49\linewidth]{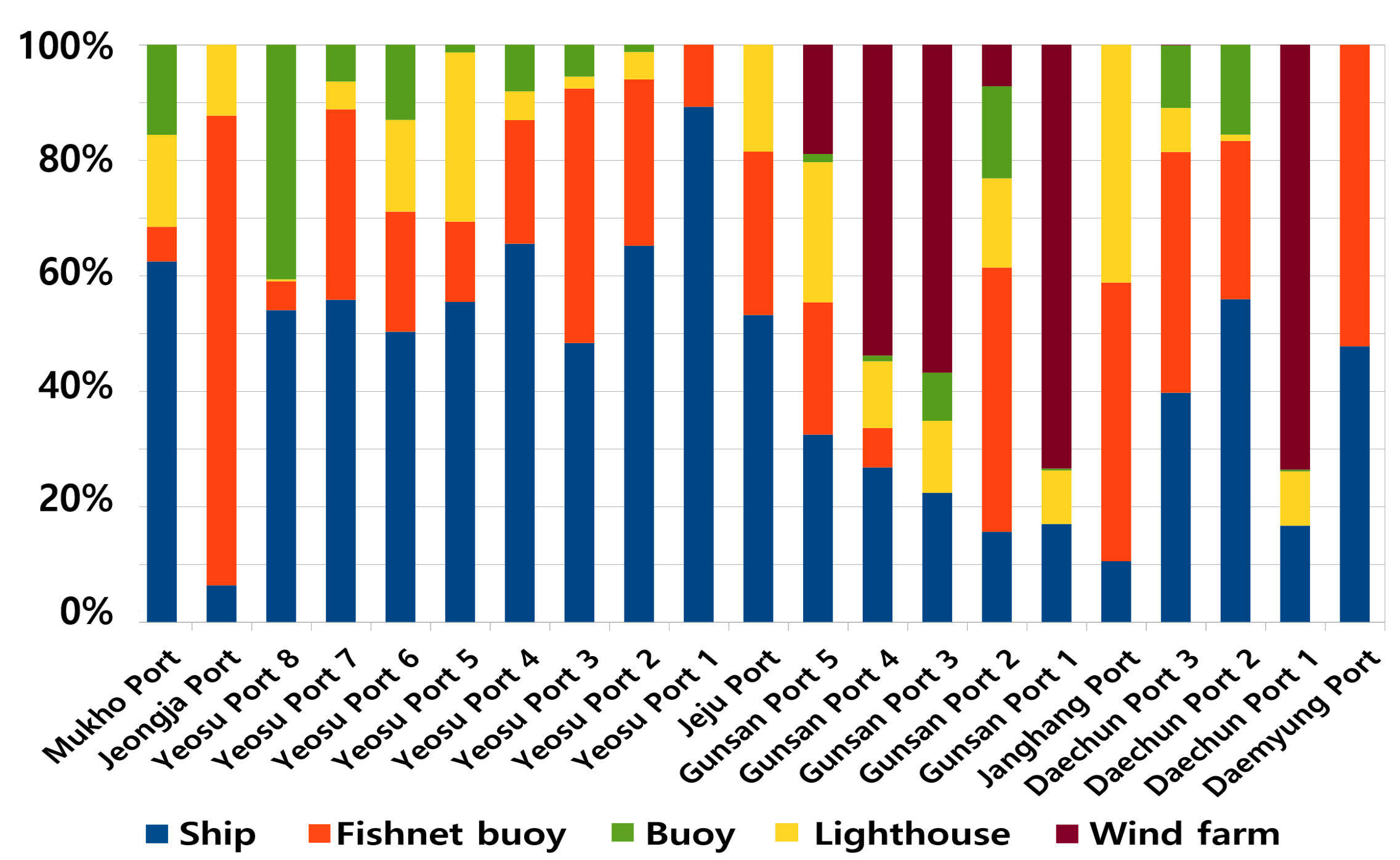}
  \caption{\textbf{Left:} Bounding boxes shape distribution for KOLOMVERSE. \textbf{Right:} Class distribution for each territorial water of KOLOMVERSE.}
  \label{Bounding}
\end{figure}

Moreover, visualization of bounding boxes shape distribution and class distribution for each territorial water of KOLOMVERSE is represented in Figure~\ref{Bounding}.

\section{Experimental results and discussion}\label{section 5}

\subsection{Implementation details}
The YOLO object detection series \cite{redmon2016you, redmon2017yolo9000, redmon2018yolov3, bochkovskiy2020yolov4, jocher2022ultralytics, wang2022yolov7} is a one-stage object detection model series that has versions ranging from YOLOv1 up to v7 and a few other variations with high accuracy and speed in the benchmark datasets. Using GPU, YOLO can be trained as a super fast and accurate object detector. Therefore, we show our data efficiency in the newer versions of YOLO, namely YOLOv3 \cite{redmon2018yolov3}, v4 \cite{bochkovskiy2020yolov4}, v5 \cite{jocher2022ultralytics}, v7 \cite{wang2022yolov7} and v8 \cite{YOLOV8S}. We trained our dataset KOLOMVERSE with the YOLOv3 and YOLOv4 models using this repository\footnote{https://github.com/AlexeyAB/darknet} and YOLOv5 model with this repository \footnote{https://github.com/ultralytics/yolov5}. Moreover, we trained YOLOv7 on our dataset from the repository \footnote{https://github.com/WongKinYiu/yolov7} and YOLOv8 from the repository \footnote{https://docs.ultralytics.com/modes/train/}. Huang et al. \cite{huang2017speed} released the TensorFlow Object Detection API that supports TenforFlow2(TF2) for fast implementation of object detectors. The API contains codes for training models such as, SSD \cite{liu2016ssd} with MobileNet \cite{howard2017mobilenets}, RetinaNet \cite{Lin_2017_ICCV}, FRCNN \cite{ren2015faster}, Mask R-CNN \cite{he2017mask} as well as SSD with EfficientDet \cite{tan2020efficientdet} and CenterNet \cite{zhou2019objects}. We trained SSD, CenterNet, and FRCNN on the KOLOMVERSE using this repository\footnote{https://github.com/tensorflow/models/tree/master/research/object$\_$detection}. SSD is a single shot multibox detector that eliminates proposal generation and computes everything in a single network \cite{liu2016ssd} and EfficientDet is a new family of object detectors, which consistently achieve much better efficiency than prior architectures \cite{tan2020efficientdet}. We performed experiments on SSD with EfficientDet architecture as the backbone. CenterNet is a simple and effective anchor-free architecture that considers an object as the center point of its bounding box. It is differentiable, faster, and better than other similar detectors \cite{zhou2019objects}. We performed experiments on CenterNet with Resnet101 architecture \cite{he2016deep} as backbone. The FRCNN \cite{ren2015faster} is a two-stage architecture and a state-of-the-art object detection tool based on a region proposal along with a fast R-CNN \cite{girshick2015fast} network. Two-stage architectures slide a window through a feature map to localize objects, classify them and regress bounding boxes. In this study, we performed experiments on our dataset on the FRCNN network with Resnet50 \cite{he2016deep} and InceptionResNetV2 \cite{szegedy2017inception} as backbone.

\subsection{Experimental settings}
To train YOLO models with our dataset, we first converted our dataset into the YOLO format. We did not change much of the pre-trained architectures except that we modified the last fully connected layers from MS COCO's 80 classes \cite{lin2014microsoft} to 5 classes for our 5-class dataset as the pre-trained models were trained on the MS COCO dataset. Additionally, we converted the configuration file of YOLO to be trained on our five class datasets. To train the YOLOv3 and YOLOv4 architectures, we resized the input size to 416$\times$416 as this is the default input size of these architectures. For the newer versions of YOLO such as, YOLOv5, YOLOv7 and YOLOv8, we use the default input size 640$\times$640. Similar to the YOLO preprocessing, we firstly converted our dataset to train it on the FRCNN model. We resized the input size to 640$\times$640 and 1024$\times$1024 and finetuned the FRCNN with a momentum optimizer and a cosine decay learning rate of base 0.0004. For the CenterNet model, we resized the images to 512$\times$512 and 1024$\times$1024 and finetuned the model with ADAM optimizer and a cosine decay learning rate of base 1$e$-6. Furthermore, for training SSD with EfficientDet$-$D0, images were resized to input size 512$\times$512, for training with backbone EfficientDet$-$D1, images were resized to 640$\times$640 and for training SSD with EfficientDet$-$D4, images were resized to input size 1024$\times$1024. The SSD models were finetuned with a momentum optimizer and a cosine decay learning rate of base 8$e$-4 for SSD with backbone EfficientDet$-$D0, 8$e$-5 for SSD with backbone EfficientDet$-$D1 and 8$e$-2 for SSD with backbone EfficientDet$-$D4. In addition, MaskRCNN was trained with a momentum optimizer and a learning rate of 0.008. We apply early stopping for the models trained with TF2 to avoid overfitting. After training, we performed evaluations on both our validation and our test set. We performed our experiments using Tensorflow (YOLOv3, YOLOv4), PyTorch (YOLOv5, YOLOv7 and YOLOv8), and TF2 (SSD, MaskRCNN, CenterNet, and FRCNN) and deployed an NVIDIA RTX 6000 GPU with 40GB memory for the experiments. Since KOLOMVERSE has images of 4K resolution with 3840$\times$2160 size, we trained the object detectors on both high and low resolution input sizes with varied backbones and compared the results. 


\begin{table*}[ht]
\footnotesize
  \caption{{State-of-the-art detector performances mean average precision at IOU 0.5 (mAP50) and IOU 0.75 (mAP75) in the validation as well as test set with varied backbone and input sizes.}}
  \centering
  \begin{tabular}{lllllll}
    \toprule

    Model & Backbone & Input size & \multicolumn{2}{c}{mAP50($\%$)} & \multicolumn{2}{c}{mAP75($\%$)} \\ 
  \cmidrule(r){4-7}
 & & & Validation set & Test set & Validation set & Test set \\
 
    \midrule

SSD &EfficientDet$-$D0 \cite{tan2020efficientdet}&512$\times$512&46.61&46.64  &14.37&14.67\\
SSD &EfficientDet$-$D1 \cite{tan2020efficientdet}&640$\times$640&52.11 &52.09  &19.30&19.57\\
YOLOv3 & Darknet53 \cite{redmon2018yolov3} &416$\times$416&59.87  & 59.57 &15.59&15.58\\
YOLOv4 & CSPDarknet53 \cite{bochkovskiy2020yolov4} &416$\times$416&60.51 & 59.99 &20.00&19.72\\
YOLOv5&
FocusCSPDarknet53\cite{jocher2022ultralytics}&
640$\times$640&
85.90&
84.80&
50.00&
49.80\\

YOLOv7&E$-$ELAN \cite{wang2022yolov7}&640$\times$640&82.20&81.20&45.70&45.60\\
YOLOv8&
YOLOv8s\cite{YOLOV8S}&
640$\times$640&
84.33&
84.67&
62.33&
62.68\\

FRCNN &Resnet50 \cite{he2016deep}&640$\times$640& 60.85 & 61.28 &30.16&30.45\\
FRCNN &InceptionResnetV2  \cite{szegedy2017inception}&640$\times$640& 55.35   &57.01 & 9.124 &      10.61 \\
CenterNet &Resnet101 \cite{he2016deep}&512$\times$512&61.40 &61.86&21.98& 22.38 \\
SSD & EfficientDet$-$D4 \cite{tan2020efficientdet}&1024$\times$1024 &61.57 &62.00&24.67& 25.26\\
FRCNN &InceptionResnetV2 \cite{szegedy2017inception} & 1024$\times$1024 & 76.05 & 76.52    &19.66 &19.61  \\
CenterNet&Resnet101 \cite{he2016deep}&1024$\times$1024 &71.24&71.32&29.51&29.93\\
MaskRCNN &InceptionResnetV2 \cite{szegedy2017inception}&1024$\times$1024 & 81.56  &81.82    &40.87  &41.61\\
    \bottomrule
    \label{Tbl2}
  \end{tabular}
\end{table*}

\begin{table*}[ht]
\footnotesize
  \caption{Class-wise experimental results of the object detectors trained on the KOLOMVERSE.}
 \centering
  \begin{tabular}{llllllll}
    \toprule
    
Model & Input size & Evaluation Set  & \multicolumn{5}{c}{mAP50($\%$)} \\ 
  \cmidrule(r){4-8}
 & & &Ship & Buoy & Fishnet Buoy & Lighthouse & Wind Farm \\

\midrule
SSD&512$\times$512&val& 69.03& 49.48 & 25.76 & 47.57 & 41.21 \\
(EfficientDet$-$D0)&512$\times$512 &test& 68.60& 49.25 & 25.67& 47.87 &41.83\\
SSD&640$\times$640&val&74.15 & 56.27 & 28.82 & 51.60 &49.73 \\
(EfficientDet$-$D1)&640$\times$640& test&73.95 & 55.13 & 28.80 & 52.36 &50.22\\
YOLOv3&416$\times$416&val&78.08 & 64.70 & 35.04 & 64.14 &57.40\\
(Darknet53)& 416$\times$416&test& 77.35& 63.59 &34.61 &64.71 &57.55\\
YOLOv4&416$\times$416&val&78.74 & 64.93& 38.08& 65.48 &55.33 \\
(CSPDarknet53)& 416$\times$416&test& 77.99&63.84 & 36.13&66.16 &55.84\\
YOLOv5&640$\times$640&val&90.10&90.90&77.60&89.10&81.60\\
(FocusCSPDarknet53)&640$\times$640&test&89.70&88.90&74.50&89.50&81.50\\
YOLOv7&640$\times$640&val&88.00&86.00&69.20&85.50&82.20\\
5(E$-$ELAN)&640$\times$640&test&87.60&83.80&66.50&86.20&81.90\\
YOLOv8&640$\times$640&val&93.70&79.70&74.40&85.50&88.40\\
(YOLOv8s)&640$\times$640&test&93.70&80.00&75.50&86.20&88.00\\
FRCNN&640$\times$640&val&82.83 & 59.21 & 46.41 & 59.08 & 56.71 \\
(Resnet50)& 640$\times$640&test&82.59 &58.16 & 46.76 &61.04 &57.85\\
FRCNN&640$\times$640&val&75.92&50.10& 50.89& 52.39& 47.44\\
(InceptionResnetV2)&640$\times$640&test& 74.69 & 59.90 & 36.10 & 54.49 & 59.90\\
CenterNet&512$\times$512&val&77.67 & 65.92 &47.44 & 64.69 & 51.26 \\
(Resnet101)&512$\times$512&test&77.34 & 65.77 & 48.04 & 65.50 &52.66\\
SSD&1024$\times$1024&val&79.70 & 50.02 & 50.33 & 70.66 &57.13 \\
(EfficientDet$-$D4)&1024$\times$1024&test&79.54 &51.43 & 49.82 & 71.29 &57.90 \\
FRCNN&1024$\times$1024&val&82.19 & 78.07 & 62.68 & 76.71 & 80.59 \\
(InceptionResnetV2)&1024$\times$1024&test&82.27&79.16&62.20 &77.80 & 81.18 \\
CenterNet&1024$\times$1024&val&69.52 & 77.24& 51.72& 75.42 &82.28 \\
(Resnet101)&1024$\times$1024&test&69.28 & 77.41 & 51.25 & 76.19 & 82.48\\
MaskRCNN&1024$\times$1024 &val& 83.30 & 82.98 &69.19 &86.13 & 86.19 \\
(InceptionResnetV2)&1024$\times$1024&test&83.37&83.47&69.14&86.96& 86.17\\

  \bottomrule
  \label{Tbl3}
  \end{tabular}
\end{table*}

\subsection{Experimental results}

Table~\ref{Tbl2} shows the evaluation results of the state-of-the-art detectors when trained and evaluated on our dataset. The results showed mean average precision (mAP) at an intersection over union (IOU) 0.50 and 0.75 for both the validation (val) and the test sets. Furthermore, in Table~\ref{Tbl3}, we present a class-wise performance analysis for both the validation and the test set. We observe that the ship class has the best performance among all the five classes because it has relatively more instances. 
As the samples of fishnet buoy are very small, it has the least mAP among all the 5 classes. However, the detectors trained with high resolution images show better performance in the fishnet buoy class. 




\begin{figure}[ht]
\centering
\includegraphics[width=0.5\linewidth,height=4.1cm]{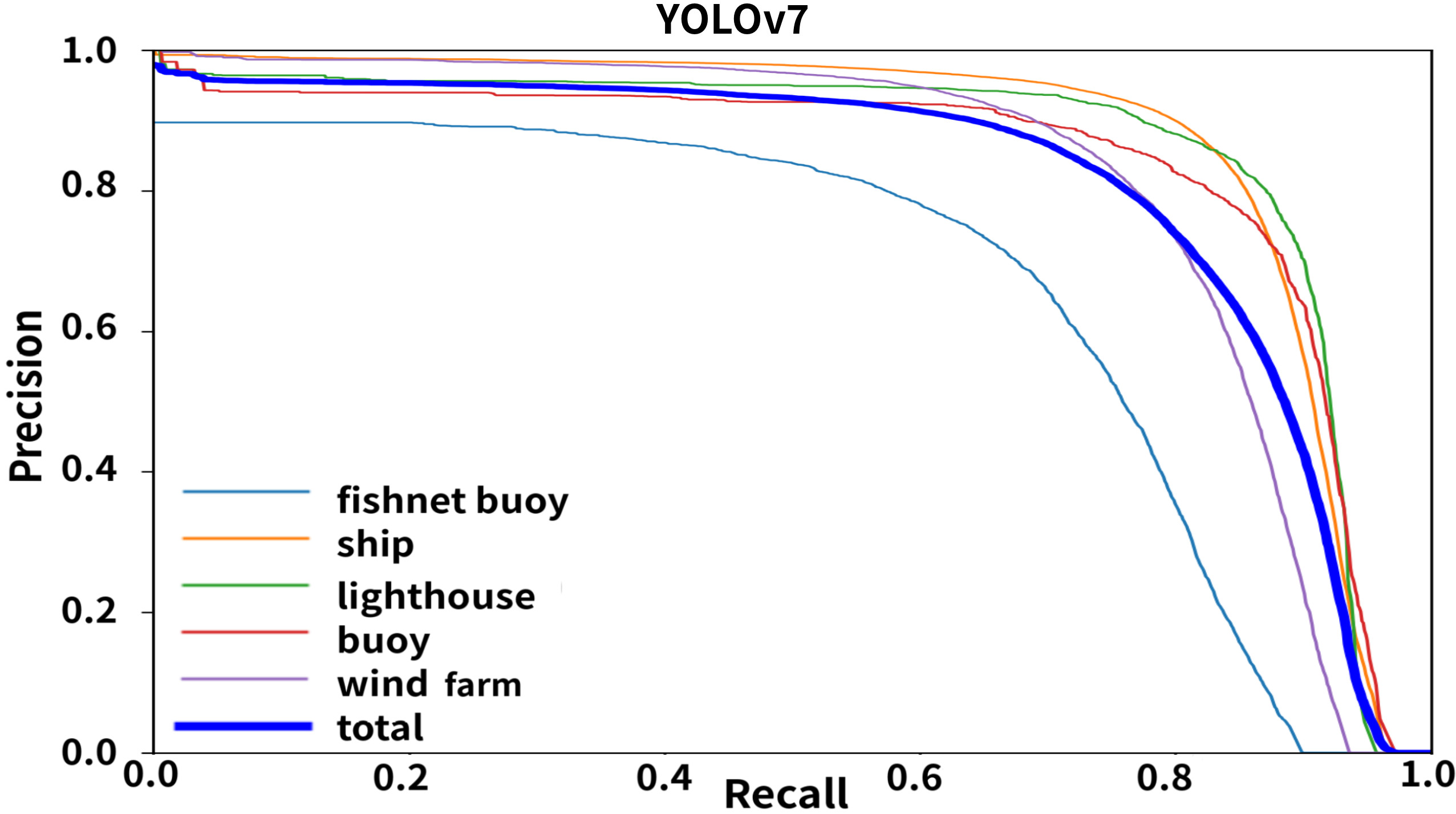}
\includegraphics[width=0.49\linewidth,height=4.1cm]{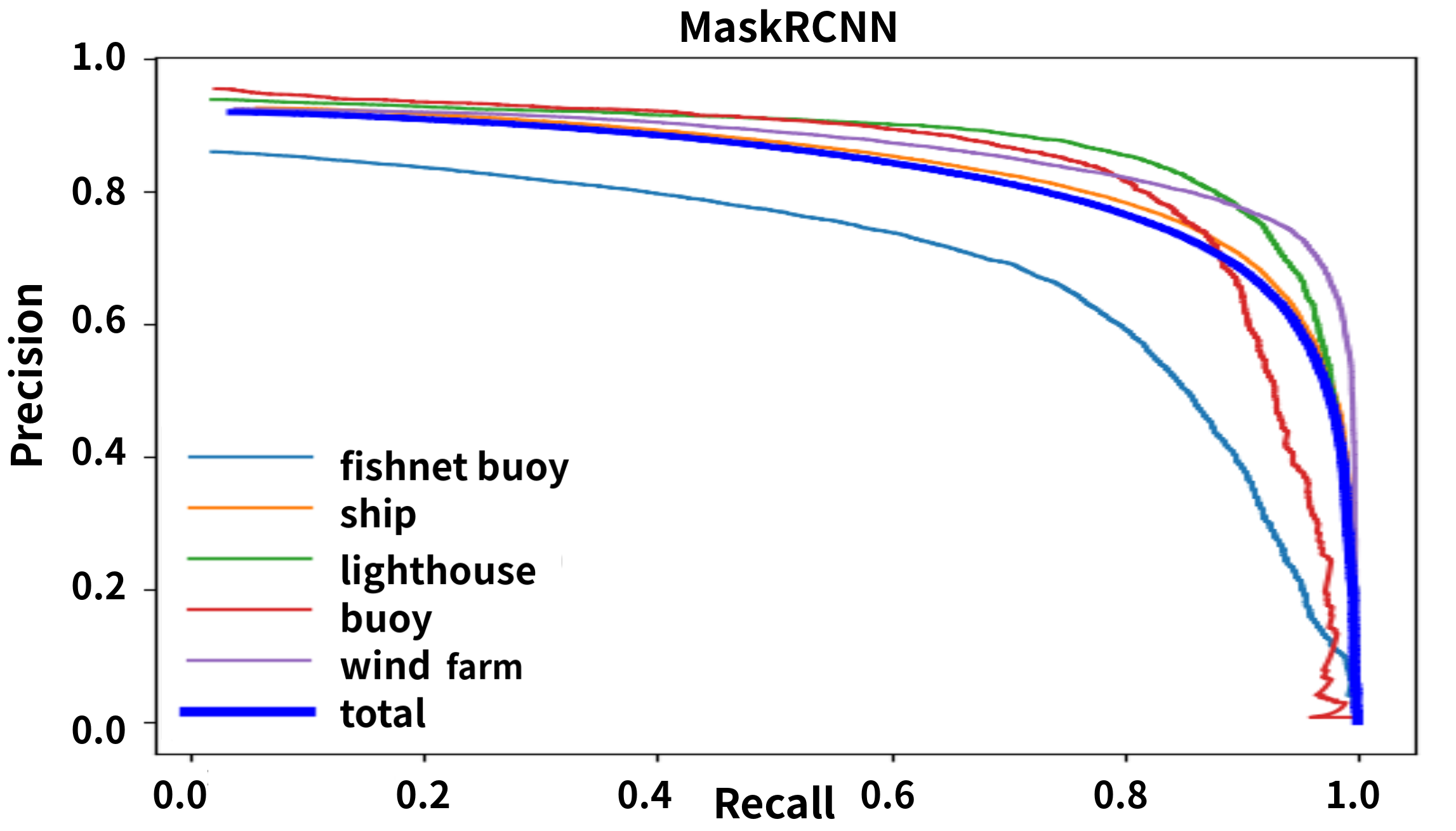}
\includegraphics[width=0.49\linewidth,height=4.1cm]{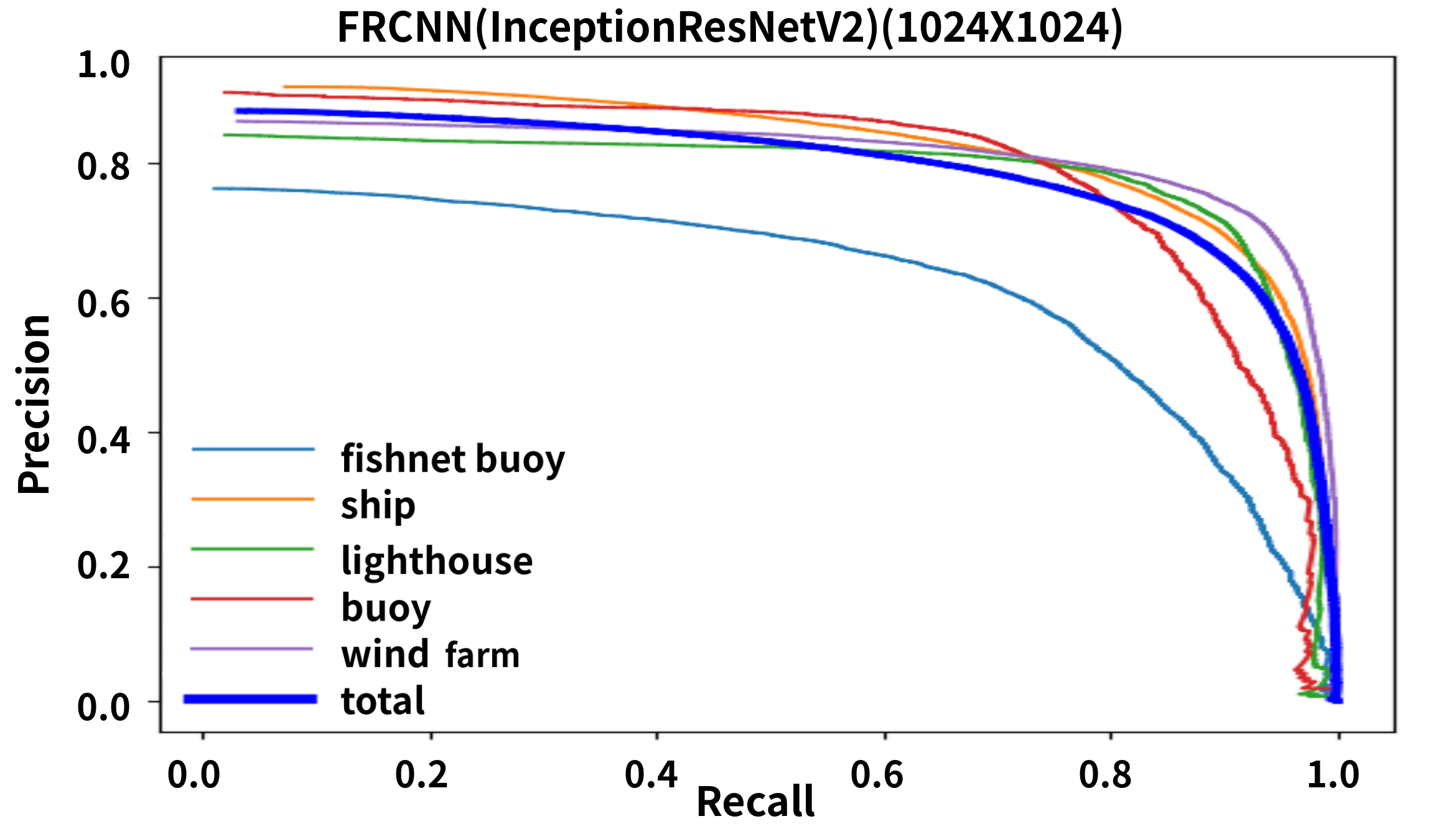}
\caption{Class-wise PR curves of three well performing detection models YOLOv7, MaskRCNN and FRCNN (InceptionV2) trained and evaluated on the KOLOMVERSE.}
\label{Fig20}
\end{figure}

\begin{figure}[ht]
\centering
\includegraphics[width=0.325\linewidth]{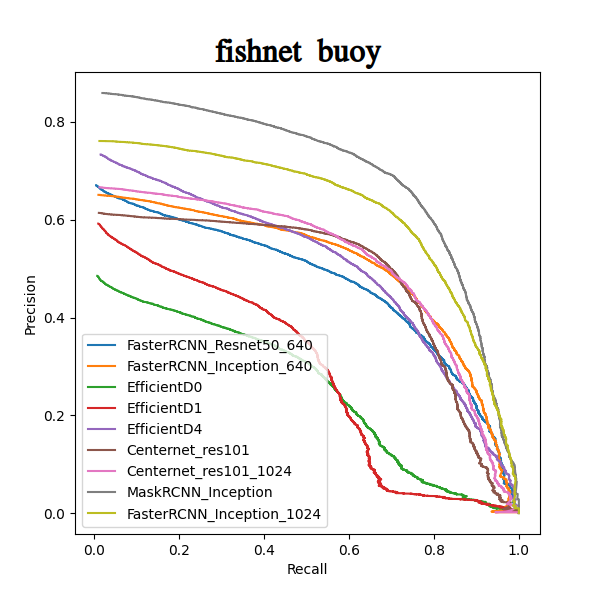}
\includegraphics[width=0.325\linewidth]{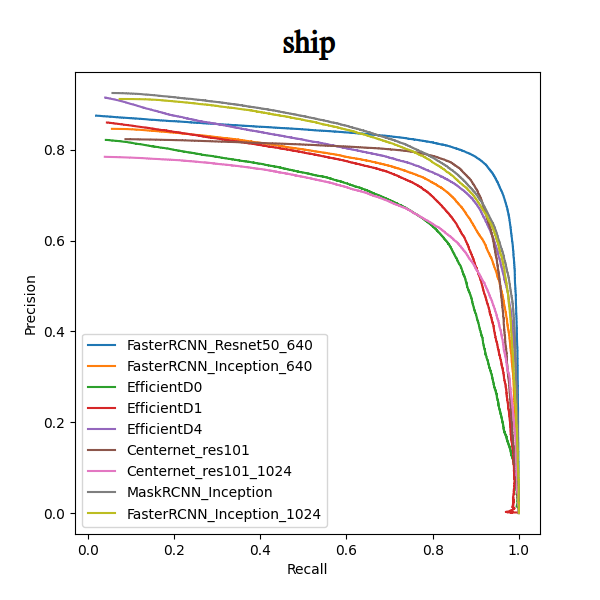}
\includegraphics[width=0.325\linewidth]{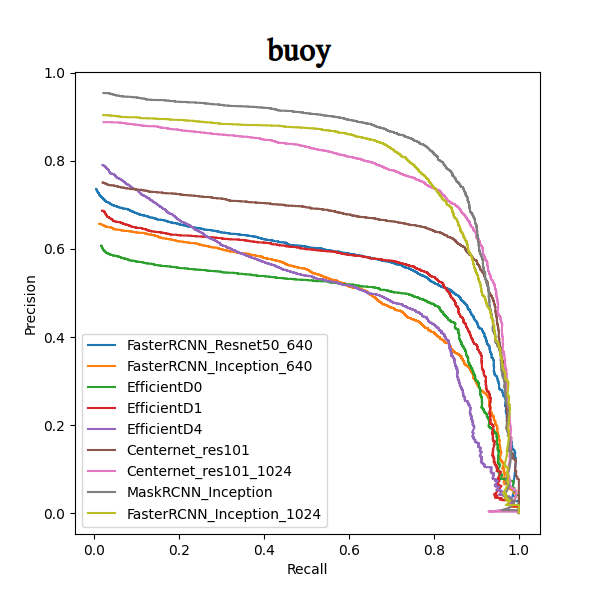}
\includegraphics[width=0.325\linewidth]{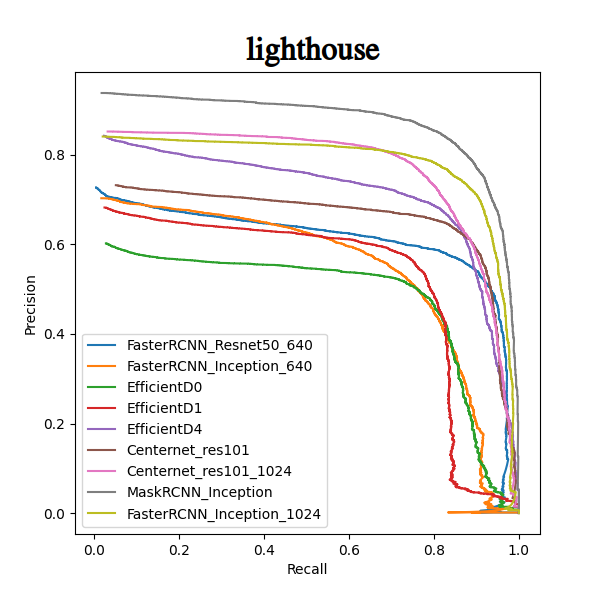}
\includegraphics[width=0.325\linewidth]{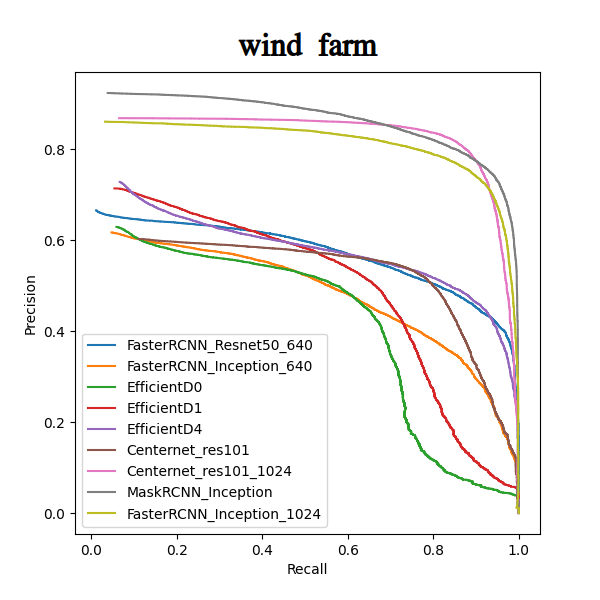}
\caption{Precision-Recall curves of object detectors trained in KOLOMVERSE on the 5 classes.}
\label{PR-curvestotal}
\end{figure}

We also present the class-wise precision-recall (PR) curves of the three well performing models trained on the KOLOMVERSE. Figure~\ref{Fig20} shows the PR curves of YOLOv7, MaskRCNN, and FRCNN (InceptionResnetV2) models trained on the KOLOMVERSE. In addition, we show the detection results of SMD, MarDCT, Seaships and KOLOMVERSE datasets in terms of mAP at IOU 0.50 in Table~\ref{Tbl5}. We only considered the ship classes of these datasets for a fair comparison. We collected ship classes from all the datasets and split them into train and test sets in the ratio of 8:2. Using Yolov8s model, we trained the datasets for a maximum of 100 epochs and maintained the same experimental conditions for all. The number of ship samples in each dataset are SMD (train : 15960, test: 3989), 
MarDCT (train: 1,939, test: 484), Seaships (train:5,600, test: 1,400) and KOLOMVERSE (train:315,149, test: 78,787).
We evaluated the trained models on the ship class samples of all the datasets. From the experiments as shown in Table ~\ref{Tbl5}, we observe that even though KOLOMVERSE detection on own dataset (94.43$\%$) is lesser compared to the others (99.28$\%$,99.50$\%$ and 99.30$\%$), it performs better when tested in other datasets. Furthermore, the other datasets donot perform well in KOLOMVERSE test set as they lack very small ship samples gathered in high resolution. It is noteworthy that MarDCT and SMD perform poorly than KOLOMVERSE when evaluated with Seaships as Seaships are known to have ship samples of various size and shapes, similar to that of KOLOMVERSE. In addition, KOLOMVERSE performs better in Seaships test set compared to Seaships's performance in KOLOMVERSE test set. From the experiments, we observe that the KOLOMVERSE dataset have stable results in SMD, MarDCT, and Seaships, even though these datasets were all captured in different environments. Also, it is evident that the KOLOMVERSE dataset can be used to generate a reliable pretrained maritime model for various maritime detection tasks because it contains large number of images captured in high resolution with diverse size and shapes of target objects. 

\begin{figure}[bt]
\centering
\includegraphics[width=1\linewidth,height=7cm]{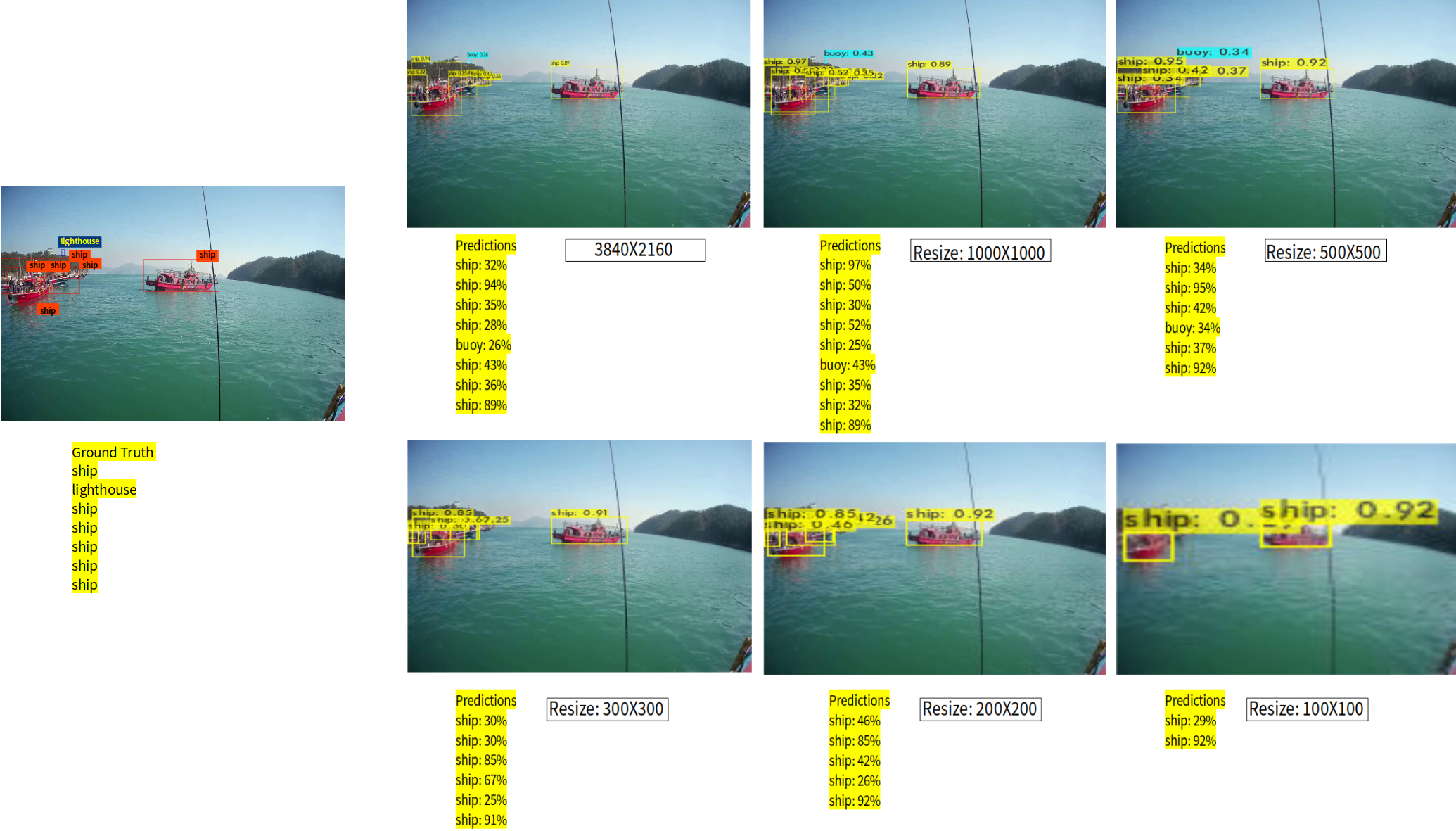}
\caption{Object detector performance in various degrees of high to low resolution images.}
\label{resolution}
\end{figure}

In addition, in Figure~\ref{PR-curvestotal} we show the Precision-Recall curves of the object detectors trained
on KOLOMVERSE for the 5 classes. The detectors fail to detect very small and heavily occluded objects
in these images. Moreover, Figure~\ref{resolution} shows how object detector performance varies in different degrees of high
to low resolution images. We observe that models trained with the higher resolution achieved better predictions, thus showing the advantages of a high resolution dataset. Furthermore, from the figure, it is evident that high resolution images are crucial to
object detection in the maritime domain.

\begin{table}[ht]
\caption{Comparison of detection results of SMD, MarDCT, Seaships and KOLOMVERSE datasets on the ship class.}
 \label{Tbl5}
 \centering
 \begin{tabular}{ccccc}
   \toprule

Train set  & \multicolumn{4}{c}{mAP50($\%$)} \\ 
\cmidrule(r){2-5}
& SMD & MarDCT & Seaships & KOLOMVERSE \\ 

\midrule

SMD&99.28&65.80&4.642&9.640\\
MarDCT &27.01&99.50&0.02254&0.3212\\
Seaships &38.80&0.2314&99.30&19.38\\
KOLOMVERSE&34.14&25.11&26.60&94.43\\

    \bottomrule
  \end{tabular}
\end{table}

\begin{figure}[ht]
\centering
\includegraphics[width=1\linewidth,height=4cm]{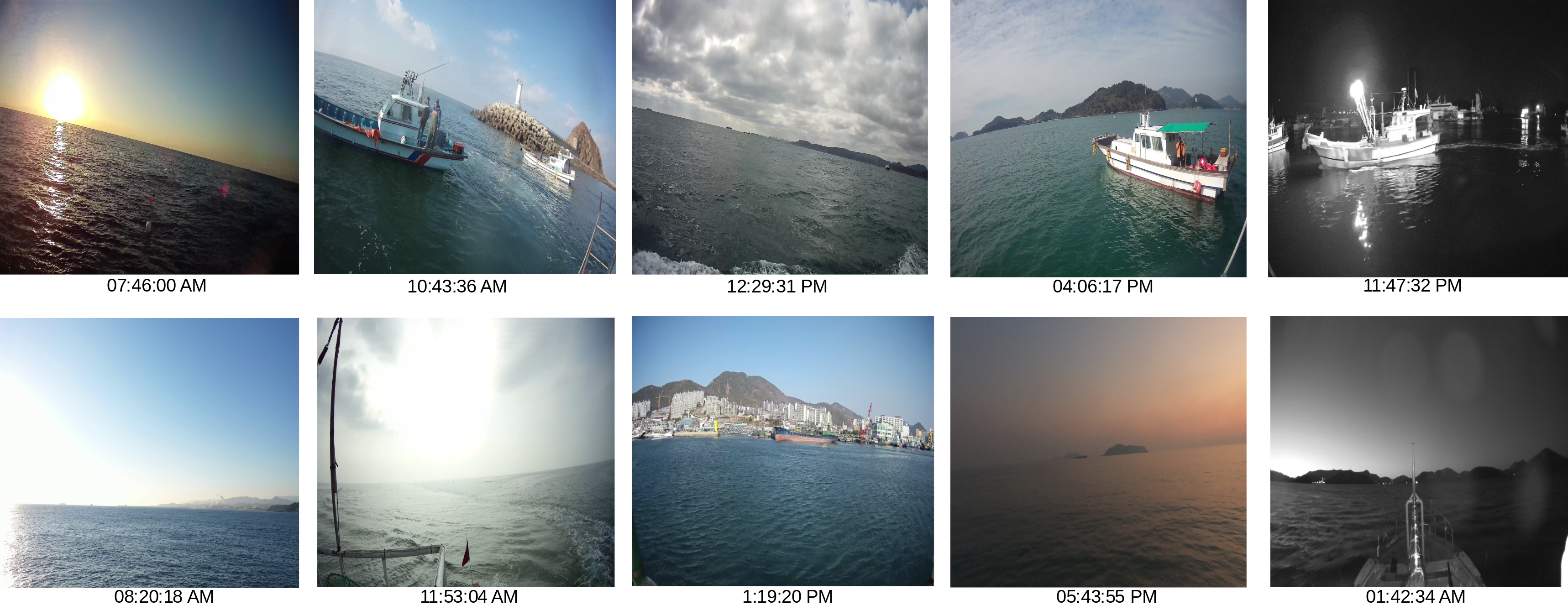}
\caption{Image samples showing variation in illumination.}
\label{Fig6}
\end{figure}

\begin{figure}[!ht]
\centering
\includegraphics[width=1\linewidth,height=6cm]{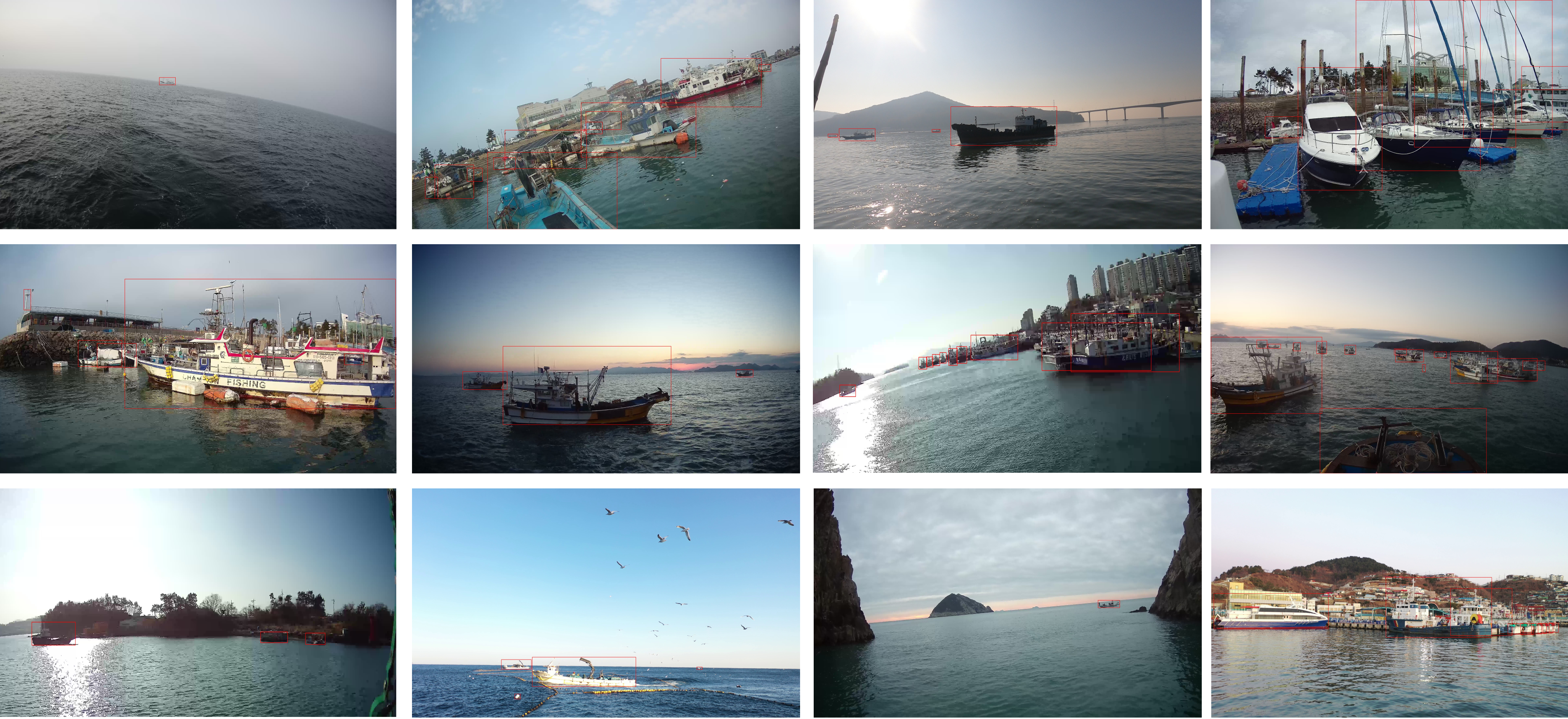}
\caption{Ship image samples showing variations in background.}
\label{Fig7}
\end{figure}

\begin{figure}[bt]
\centering
\includegraphics[width=1\linewidth,height=3cm]{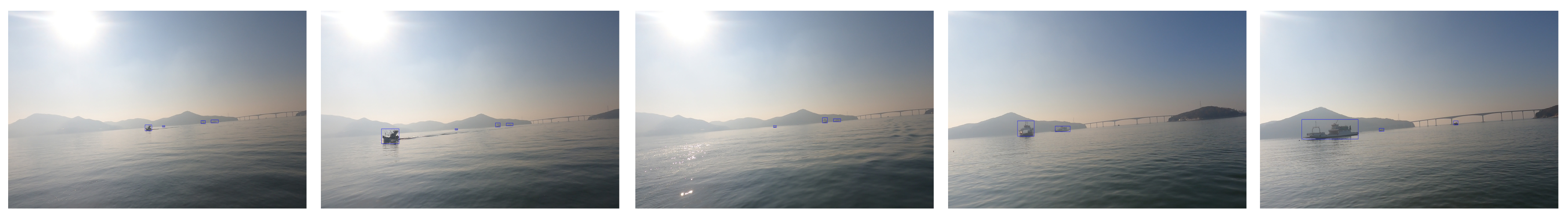}
\caption{Image samples showing variations in scale within same ship objects.}
\label{Fig8}
\end{figure}

\begin{figure}[bt]
\centering
\includegraphics[width=0.9\linewidth,height=4cm]{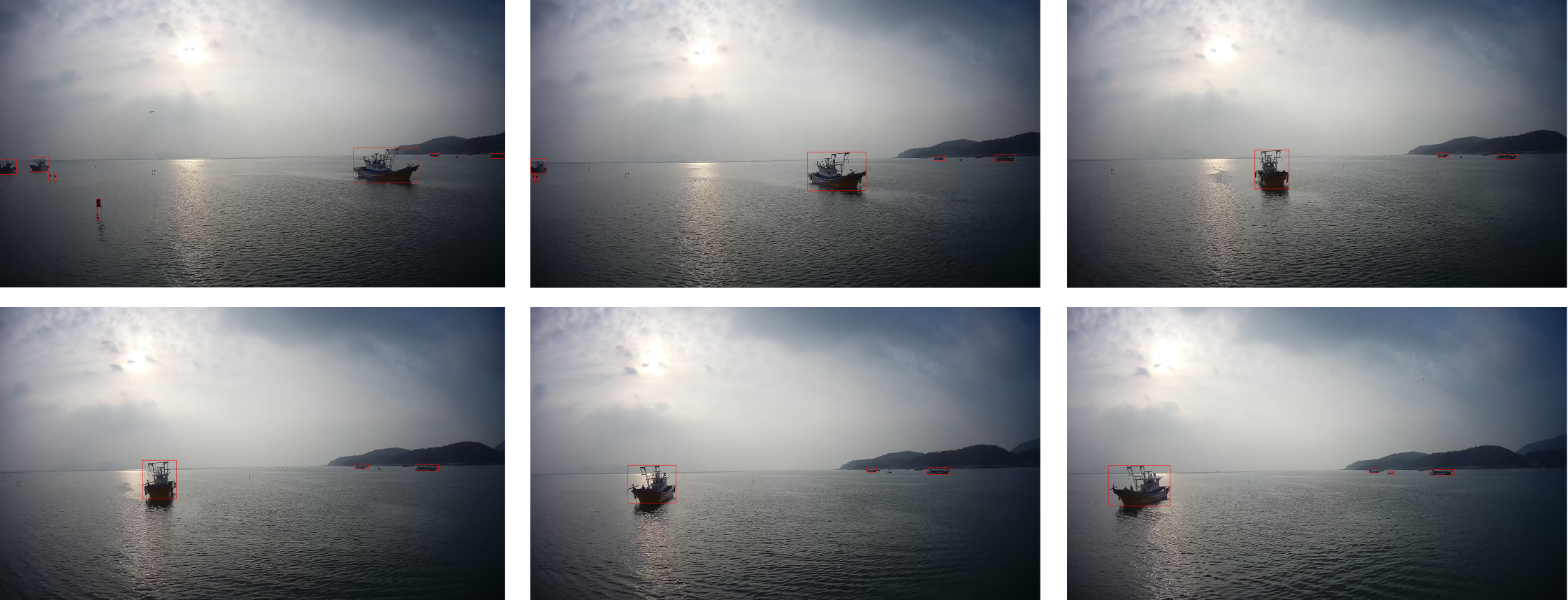}
\caption{Image samples showing variations in viewpoint in the same ship object. The viewpoint transitions from right to left.}
\label{Fig10}
\end{figure}

\begin{figure}[h!]
\centering
\includegraphics[width=0.9\linewidth,height=4cm]{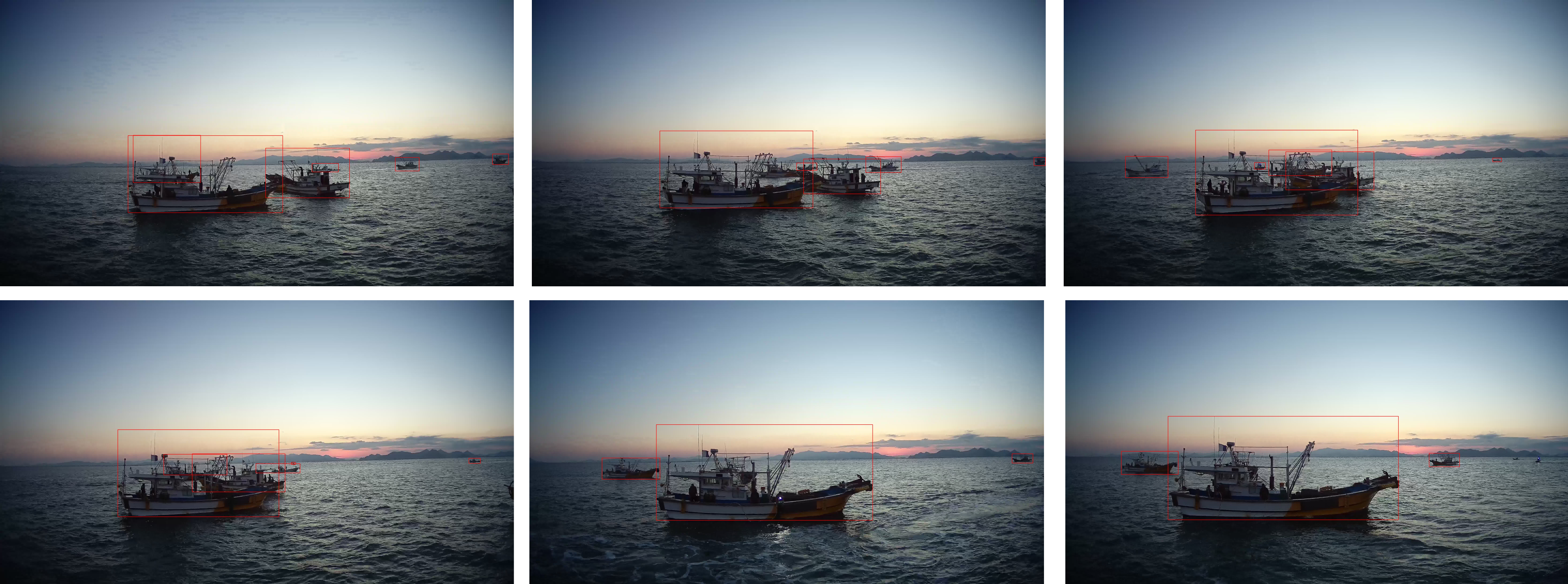}
\caption{Image samples showing varying levels of occlusion among the same ship objects.}
\label{Fig11}
\end{figure}

\begin{figure}[h!]
\centering
\includegraphics[width=0.9\linewidth,height=4cm]{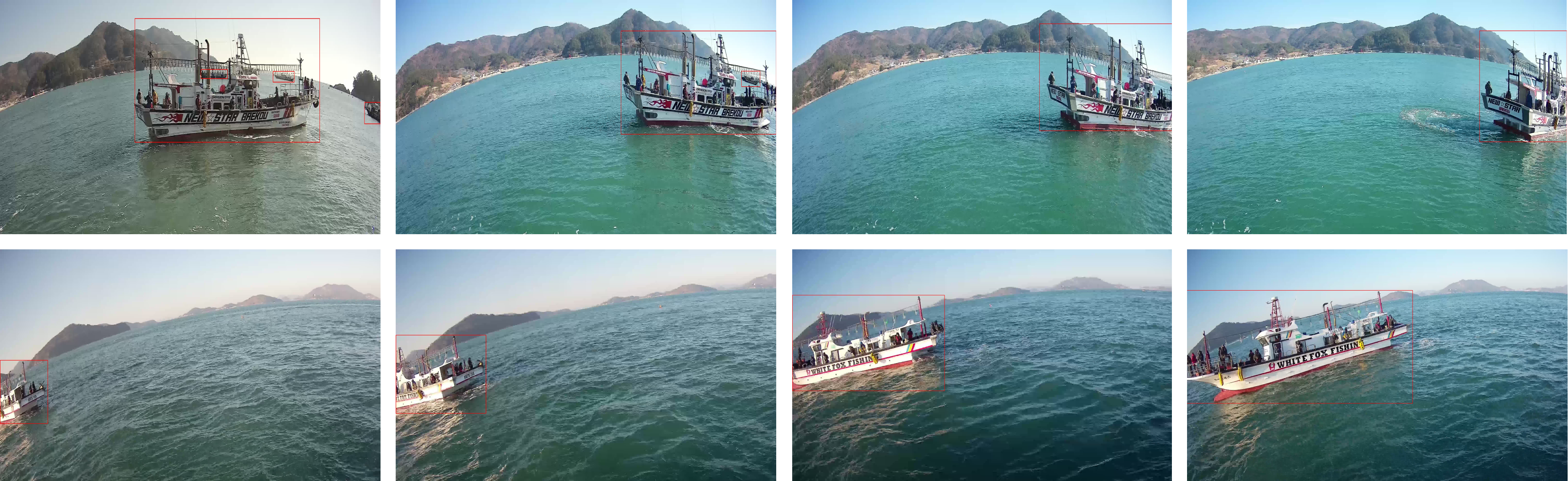}
\caption{Image samples showing variations in proportions in the same ship objects.}
\label{Fig12}
\end{figure}

KOLOMVERSE has variations in illumination, viewpoint, occlusion, background, scale and proportion that makes object detectors trained with the KOLOMVERSE more suitable for real-time applications.

Figure~\ref{Fig6} shows image samples with illumination variation.
Figure~\ref{Fig7} shows image samples with background variation.
Figure~\ref{Fig8} shows image samples with scale variation in ship objects.
Figure~\ref{Fig10} shows image samples with viewpoint variation.
Figure~\ref{Fig11} shows image samples with occlusion variation.
Figure~\ref{Fig12} shows image samples with proportion variation.

\section{Conclusion}
We introduced a publicly available large-scale object detection dataset in the maritime domain. Our goal with this dataset is to address the scarcity of large-scale datasets for object detection in the maritime domain. The dataset comprises 186,419 4K images annotated with five maritime object classes for detection. In comparison to other datasets, our dataset includes a broader range of diverse and challenging categories essential for effective maritime object detection. We conducted training and evaluation using state-of-the-art pre-trained object detectors, achieving high accuracy in our experiments. Specifically, our dataset demonstrated superior performance in two-stage detectors.

There is still room for improvement in this research. The KOLOMVERSE was constructed using images of objects detected during ship voyages and includes objects most commonly seen at sea. 
However, because the data were collected during ship navigation, there is an imbalance with more instances of ships compared to other categories. Therefore, it is essential to address this data imbalance by collecting more data for other categories in future efforts. Additionally, the KOLOMVERSE predominantly features fishing boats. We aim to rectify this by incorporating various types of ship objects in the future.

\section*{Acknowledgment}
The authors are grateful to the anonymous reviewers for reading the manuscript carefully and providing constructive comments which greatly helped to improve this paper. In addition, this research was supported by Korea Research Institute of Ships and Ocean engineering a grant from Endowment Project of “Development of Open Platform Technologies for Smart Maritime Safety and Industries” funded by Ministry of Oceans and Fisheries(2520000292, PES5230), and also supported by the National Information Society Agency of "Marine Object AI Data" funded by Ministry of Science and ICT.

\ifCLASSOPTIONcaptionsoff
  \newpage
\fi



\bibliographystyle{IEEEtran}
%

\bibliography{IEEEabrv}

\begin{IEEEbiography}[{\includegraphics[width=1in,height=1.25in,trim={3.6cm, 0, 4cm, 0}, clip,keepaspectratio]{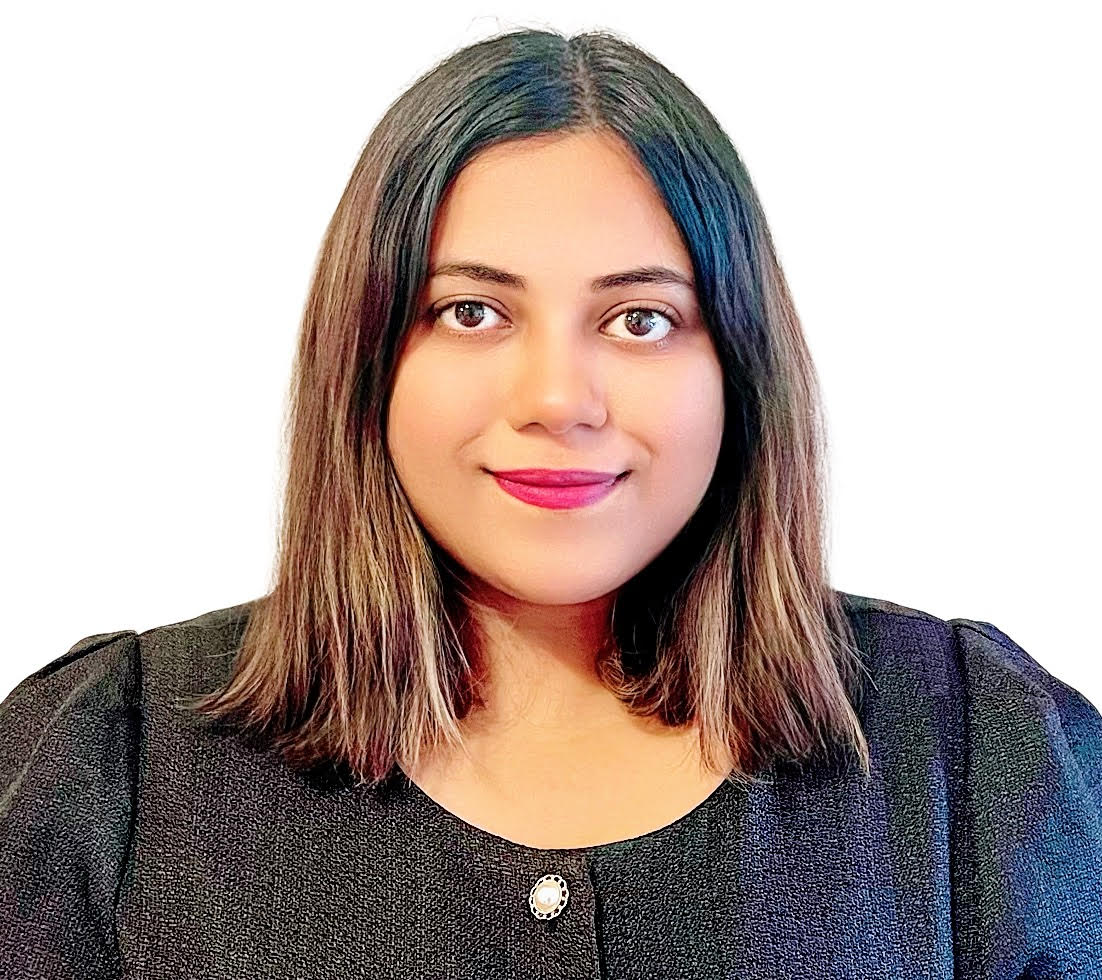}}]{Abhilasha Nanda} received the B.Tech. degree in Information Technology from Vellore Institute of Technology, Vellore, India, in 2013, the M.S. degree in Information Technology from the Hong Kong University of Science and Technology, Hong Kong, in 2014, and the Ph.D. degree in Computer Science from the Korea Advanced Institute of Science and Technology, Daejeon, Republic of Korea, in 2021.
From July 2021 to April 2023, she was a Postdoctoral Researcher at the Maritime Digital Transformation Research Center, Korea Research Institute of Ships and Ocean Engineering, Republic of Korea. From May 2023 to May 2024, she was a Senior Researcher at AIVENAUTICS (AIVN), Republic of Korea. Her research interests include computer vision, data science, and image processing.
\end{IEEEbiography}

\begin{IEEEbiography}[{\includegraphics[width=1in,height=1.25in,clip,keepaspectratio]{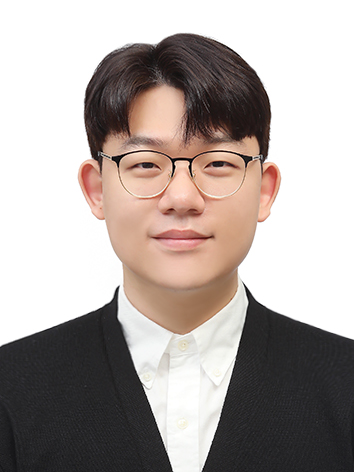}}]{Sung Won Cho} received the B.S. degree in Logistics from the Korea Aerospace University, Gyeonggi-do, Republic of Korea, in 2016, and the Ph.D. degree in Industrial and Management Engineering from Korea University, Seoul, Republic of Korea, in 2021.

From June 2019 to August 2023, he was a Senior Researcher at the Maritime Digital Transformation Research Center, Korea Research Institute of Ships and Ocean Engineering, Republic of Korea. Since 2023, he has been an Assistant Professor with the Department of Management Engineering, Dankook University. His research interests include maritime logistics, intelligent transportation systems and optimization methods.
\end{IEEEbiography}

\begin{IEEEbiography}[{\includegraphics[width=1in,height=1.25in,clip,keepaspectratio]{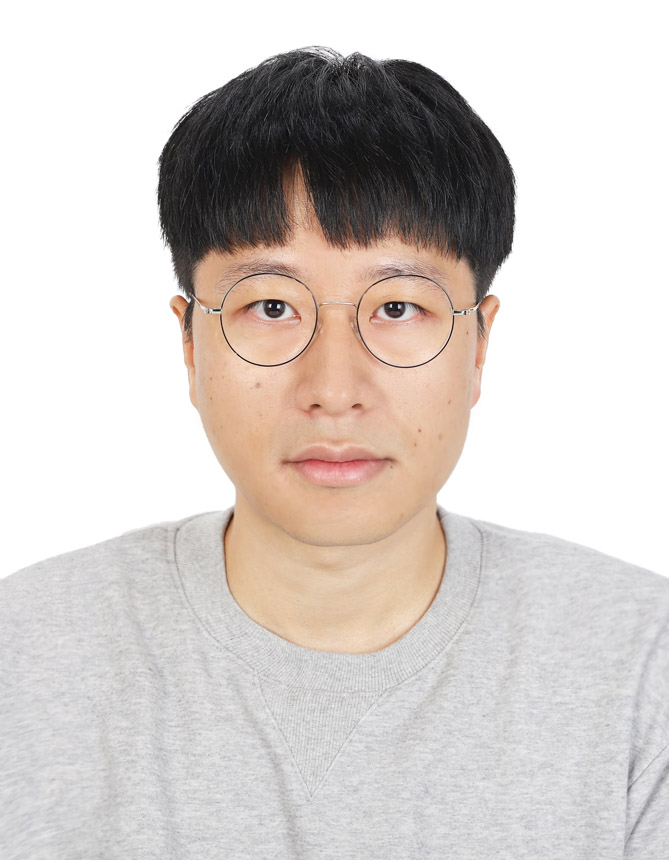}}]{Hyeopwoo Lee} received his B.S. degree in Electrical Engineering and Computer Science from the Gwangju Institute of Science and Technology, Gwangju, Republic of Korea, in 2014, and his M.S. and Ph.D. degrees in the School of Computing from the Korea Advanced Institute of Science and Technology in 2021. From 2021 to 2023, he was a postdoctoral researcher at the Maritime Digital Transformation Research Center, Korea Research Institute of Ships and Ocean Engineering, Republic of Korea. Since 2023, he has been the director of the AI Research Center at AIVENAUTICS (AIVN). He is currently working on vision image models for maritime applications and developing situation awareness solutions through sensor fusion.
\end{IEEEbiography}

\begin{IEEEbiography}[{\includegraphics[width=1in,height=1.25in,clip,keepaspectratio]{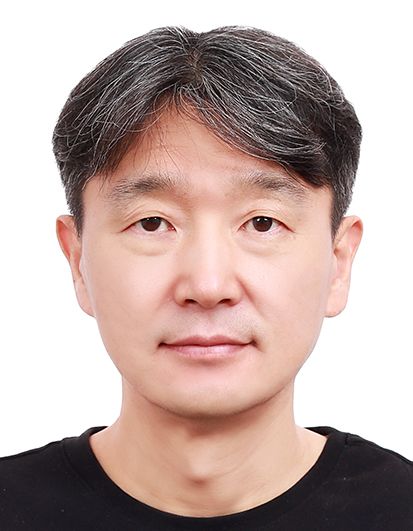}}]{Jin Hyoung Park} is the CEO and co-founder of AIVENAUTICS (AIVN). He is suspending his affiliation at the Korea Research Institute of Ships and Ocean Engineering (KRISO). AIVN is a startup company focusing on maritime autonomous navigation assistance system technologies. He received his B.S. and M.S. degrees in Computer Science from Kyungpook National University and his Ph.D. in Computer Science from Korea Advanced Institute of Science and Technology (KAIST). Before founding AIVN, he led the MAIDaS (Maritime AI $\&$ Data Science) project at KRISO as the project manager from January 2021 to December 2022. As an activity for his international standard development in maritime digitalization, he has been serving as the convenor of the ISO/IEC JTC 1/SC 41/WG 7 (Maritime, Underwater IoT, and Digital Twin Applications) since December 2022. His research interests include maritime digitalization and digital transformation technologies.
\end{IEEEbiography}

\end{document}